\RequirePackage{fix-cm}
\documentclass[twocolumn] {svjour3} 
\smartqed  
\usepackage{graphicx}
\usepackage[caption=false]{subfig}
\usepackage[numbers]{natbib}
\usepackage{amsfonts}

\renewcommand{\tilde}{\widetilde} 
\journalname{}
\begin{document}

\title{Supervised Learning in the Presence of Concept Drift
}
\subtitle{A modelling framework}


\author{M.~ Straat \and F.~Abadi \and Z.~Kan \and 
C.~G{\"o}pfert \and B.~Hammer  \and M.~Biehl }


\institute{Michiel Straat \and Zhuoyun Kan  \and Michael Biehl$^{\ast}$  \at
				Bernoulli Institute for Mathematics, Computer Science and Artificial
                Intelligence, University of Groningen, Nijenborgh 9, 9747 AG Groningen, The Netherlands\\
				\(^\ast\) Corresponding Author, 
				\email{m.biehl@rug.nl}
				\and 
				Fthi Abadi \at Aksum University, Institute of Engineering and Technology, \ Computing Science Department, Axum, Tigray, Ethiopia
				\and
				Christina G{\"opfert} \and Barbara Hammer \at
				Bielefeld University, CITEC, Machine Learning Group, 33594 Bielefeld, Germany
}


\maketitle

\begin{abstract}
We present a modelling framework for the  investigation of 
supervised learning in 
non-stationary environments.
Specifically, we model two example types of learning systems: 
prototype-based Learning Vector Quantization (LVQ) for classification 
and shallow, layered neural networks for regression tasks. 
We investigate so-called student teacher scenarios in which the systems are 
trained from a stream of high-dimensional, labeled data. 
Properties of the target
task are considered to be non-stationary due to 
drift processes while the training is
performed.  Different types of concept drift are studied,
which affect the density of example inputs only, the target rule
itself, or both. 
By applying methods from statistical physics, we develop a
modelling framework for 
the mathematical analysis of the training dynamics in non-stationary 
environments. 

Our results show that standard LVQ algorithms 
are already suitable for the training in
non-stationary environments to a certain extent. 
However, the application of \textit{weight decay} as an explicit 
mechanism of forgetting does not improve the
performance under the considered drift processes. 
Furthermore, we investigate gradient-based training of layered neural networks 
with sigmoidal activation functions and compare with the use of rectified linear units
(ReLU).  Our findings show that the sensitivity to concept drift and the effectiveness
of weight decay differs significantly between the two types of activation function.

\keywords{Classification \and regression \and supervised learning \and drifting concepts \and Learning Vector Quantization \and layered neural networks}
\end{abstract}

\sloppy 

\section{Introduction}
\label{intro}

The topic of efficiently learning from example data
in the presence of \textit{concept drift} has attracted
 significant interest in the 
machine learning community. Terms such as  
\textit{lifelong learning} or \textit{continual learning} have become
popular keywords  in this context \cite{driftreview}.

Very often, machine learning 
processes \cite{Hastie} are realized according to a standard
set-up which distinguishes two main stages:
In the first, the so-called {\it  training phase,}
parameters of the learning system
are adapted in an optimization process which
is guided by a given set of example data. 
In the following {\it working phase,} the obtained 
hypothesis, e.g.\ a classifier or regression system,
can be applied to novel data. 
This workflow relies on the
implicit assumption that the training data is indeed
representative for the target task
in the working phase. Statistical properties of the 
data and the target itself 
should not change during or after training.

However, in many practical tasks and relevant real world 
scenarios, the assumed  separation of training and working 
phase appears artificial and cannot be justified. Obviously, in
most human or other biological learning  
processes \cite{Cetraro2}, the assumption is 
unrealistic.  Similarly,  in many technical contexts,  
training data is  available as a non-stationary stream of observations.
In such settings, the separation of training and working 
phase is meaningless, 
see  \cite{Ade,Ditzler,Joshi,Losing,driftreview} for reviews. 

In the literature, 
two major types of non-stationary
environments have been discussed:
The term {\it virtual drift} refers to situations 
in which statistical properties of the training data
are time-dependent, 
 while the actual  target task remains unchanged.  
Scenarios where the target classification or regression
 scheme itself changes with time are referred to
as {\it real drift} processes.
Frequently, both effects coincide and a clear distinction
of the two cases becomes difficult. 
  
 The presence of drift
 requires some form of \textit{forgetting} of 
 dated  information while the system is adap\-ted to 
 more  recent observations. The design of 
 useful,  forgetful training schemes hinges on an 
 adequate theoretical understanding of the relevant phenomena.
 To this end, the development of a suitable modelling framework
 is  instrumental.  An overview of earlier work and more recent 
 developments  in the context of non-stationary learning
 environments can be found  in references like 
 \cite{Ade,Ditzler,Joshi,Losing,driftreview}. 
 
  Methods developed in statistical physics 
  can be applied in the 
  the mathematical description of the 
  training dynamics to obtain typical 
  \textit{learning curves}.    
  The statistical mechanics of on-line 
  learning has helped to gain insights into 
  the behavior of various learning 
  systems, see e.g.  
  \cite{handbook,engel,cambridge,revmod} and references 
  therein. 
Here, we apply these concepts to study the influence of
concept drift  and weight decay in two exemplary model situations:
prototype-based binary classification and continuous 
regression with feedforward neural networks. 
We study standard training algorithms 
under concept drift and address, both,
virtual and real drift processes.

This paper presents extensions of our contribution to the
\textit{Workshop on Self-Organizing
Maps and Learning Vector Quantization, Clustering and Visualization} (WSOM 2019)
\cite{wsom19proc}.  Consequently, parts of the text resemble or 
have been taken over literally from \cite{wsomdrift}  without explictit notice. 
This concerns, for instance, parts of the introduction and the description of 
models and methodology in Sec.\  \ref{modmet}. 
Similarly, some of the results have also been presented in  \cite{wsomdrift},
which focused  on the study of
explicitly time-dependent densities in a stream of clustered data for
LVQ training. 

{We complement our conference contribution \cite{wsomdrift} 
significantly by studying also
the influence of drift on the training of regression type layered neural
networks.
First results concerning such systems with sigmoidal
hidden unit activation function under concept drift have been published
in \cite{entropy}, recently. Here, the scope of the analysis is exten\-ded to 
layered net\-works of rectified linear units (ReLU). We 
concentrate on the  comparison 
of the latter, very popular activation function and its classical, sigmoidal counterpart
with respect to the sensitivity to drift and the effect of weight decay. }

{We have selected LVQ for classification and layered neural networks for regression
as representatives of important paradigms in machine learning. These systems
provide a workshop in which
to develop modelling techniques and analytical approaches that will facilitate
the study of other setups in the future.}

In the following section we introduce the machine learning systems, the model setup 
including the assumed densities of data, the target rules as well as the mathematical framework
of the statistical physics based analysis.  
Our results concerning classification and regression systems in the presence of concept
drift are presented and discussed in Sec.\ \ref{sec:results} before we conclude
with a summary and outlook on forthcoming investigations. 

\section{Model and Methods} \label{modmet}
In Sec.\ \ref{LVQ} we introduce Learning Vector 
Quantization for classification tasks with emphasis on the
well established LVQ1 training scheme.  We also propose a 
model density of data which was previously
investigated in the mathematical  analysis of LVQ training in 
stationary and specific non-stationary environments. 
Here, we extend the approach to 
the presence of  virtual concept drift and consider
{\it weight decay} as an explicit mechanism of
forgetting. 

Thereafter, Section \ref{SCM} presents a 
student teacher scenario
for the learning of a regression scheme with shallow,
layered neural networks of the feedforward type. 
Emphasis is on the comparison of two important types
of hidden unit activations; 
 traditional sigmoidal transfer functions and the popular
rectified linear unit (ReLU) activation. We consider
gradient-based training in the presence of 
real concept drift and also introduce weight decay as 
a mechanism of forgetting. 

A unified description of the theoretical approach
to analyse the training dynamics in classsification and regression
systems is given in Sec.\ \ref{mathdynamics}.

\subsection{Learning Vector Quantization} \label{LVQ} 
The family of LVQ 
algorithms is widely 
used for practical classification problems
\cite{wires,kohonen1,kohonen2,nova}. 
The popularity of LVQ is due to a number 
of attractive features:
It is straightforward to implement, very
flexible and intuitive. Moreover, it constitutes
a natural tool for multi-class problems. 
The actual classification scheme is 
very often based on Euclidean metrics
or other simple measures, which 
quantify the distance
of inputs or feature vectors
from the class-specific prototypes. 
Unlike many other methods,
LVQ facilitates direct 
interpretation of the classifier because 
prototypes are defined in the 
same space as the data \cite{wires,nova}. 
The approach is based on the idea of
representing classes by more or less typical
representatives of the training instances. This suggests
that LVQ algorithms should also be capable of 
tracking changes in the density of samples, a hypothesis that
has been studied for instance in \cite{wsomdrift,wsomschleif},
recently. 

\subsubsection{Nearest Prototype Classifier}
In general, several prototypes can be employed to represent each class. 
However, we restrict the analysis to 
the simple case of 
only one prototype per class in binary classification problems. Hence we 
consider two prototypes $\vec{w}_k  \in I\!\!R^N$ each representing
one of the 
classes $k\in \{1,2\}.$   
Together with a distance measure
$d(\vec{w},\vec{\xi}),$
the system parameterizes  
a Nearest Prototype Classification (NPC) scheme:
 Any given  input 
$\vec{\xi} \in I\!\!R^N$ is 
assigned to the class $k=1$ if 
$d(\vec{w}_1,\vec{\xi})< d(\vec{w}_2,\vec{\xi})$ 
 and to class $2$, otherwise. In practice,
ties can be broken arbitrarily. 

A variety of distance measures have  been
used in  LVQ, enhancing the flexibility of the approach even further \cite{wires,nova}. 
This includes the conceptually interesting
use of adaptive metrics in \textit{relevance learning,}
see  \cite{wires} and references therein.  
Here, we restrict our analysis to the simple 
(squared) Euclidean measure
\begin{equation} 
\label{squaredEuclidean} 
d(\vec{w}, \vec{\xi})= (\vec{w} - \vec{\xi})^2.
\end{equation} 
 
We assume that the training procedure provides a
stream of single examples \cite{handbook}:  
At  time step $\mu \, = \, 1,2,\ldots, $ the
vector $\vec{\xi}^{\, \mu}$  is presented, together with its given class label
$\sigma^\mu=1,2$. 
Iterative on-line LVQ updates are of the 
general form \cite{jmlr,LVQwsom,performance}
\begin{eqnarray} 
 \vec{w}_k^\mu &=&  \vec{w}_k^{\mu-1} \, + \, 
 \frac{\eta}{N} \,
  \Delta \vec{w}_k^\mu  
 \mbox{~with} \nonumber \\ 
\label{generic}
 \Delta \vec{w}_k^\mu & = &  
   f_k\left[d_1^{\mu},d_2^{\mu},\sigma^\mu,\ldots\right]
 \, \left(\vec{\xi}^\mu - \vec{w}_k^{\mu-1}\right)
 \end{eqnarray} 
where $d_i^\mu = d(\vec{w}_i^{\mu-1},\vec{\xi}^\mu)$ 
and the learning rate $\eta$ is 
scaled with
the input dimension $N$.  
The precise algorithm is specified by choice of
the {\it modulation function}  $f_k[\ldots]$, 
which depends typically on the Euclidean distances of the
 data point from the current prototype positions and on 
the labels $k,\sigma^\mu=1,2$ of the prototype and training 
example, respectively. 

\subsubsection{The LVQ1 training algorithm}
A popular and intuitive
LVQ training scheme was already 
suggested by Kohonen and is known as {LVQ1} 
\cite{kohonen1,kohonen2}.  
Following the NPC concept, it
updates only the currently closest prototype in a so-called
{\it Winner-Takes-All} (WTA) scheme.  Formally,
the LVQ1 prescription for a system with 
two competing 
prototypes is given by Eq.\ (\ref{generic}) 
with
\begin{equation} \label{LVQ1f}
f_k[d_1^\mu,d_2^\mu,\sigma^\mu] \, = \Theta\left(d_{\widehat{k}}^\mu - d_{k}^\mu\right) \Psi(k,\sigma^\mu),
\end{equation}  
where
$
\widehat{k} = \left\{ \begin{array}{ll}
   2 & \mbox{if~} k=1  \\
   1 & \mbox{if~} k=2, 
\end{array} \right. 
\mbox{and~~} 
\Psi(k,\sigma)=  \left\{ \begin{array}{ll}
   +1 & \mbox{if~} k=\sigma   \\
   -1 & \mbox{else.}  \\ 
\end{array} \right.
$
\ \\[1mm]
Here, the Heaviside function $\Theta(\ldots)$ singles out the 
winning prototype
and the factor $\Psi(k,\sigma^\mu)$ 
determines 
the sign of the update: The WTA update 
according to Eq. (\ref{LVQ1f})
moves the prototype towards the presented feature vector
if it carries the same class label  $k=\sigma^\mu$. On the 
contrary, if the prototype is meant to present a different class,
its distance from the data point is increased even further. 
Note that LVQ1 cannot be interpreted as a gradient
descent procedure of a suitable cost function in a straightforward
way due to discontinuities at the class boundaries, see \cite{jmlr} for
 a discussion and references. 

Numerous variants and modifications of LVQ 
have been presented in the literature,
aiming at better convergence or classification
performance, see
\cite{jmlr,wires,kohonen1,nova}.  
Most of these modifications, however,  
retain the basic idea of attraction and 
repulsion
of the winning prototypes.

\subsubsection{Clustered Model Data}
 LVQ algorithms are most suitable for classification
 sche\-mes which reflect a given cluster structure in 
 the data. In the modelling, 
 we therefore consider a stream of random 
 input vectors $\vec{\xi} \in \mathbb{R}^N$ which are
generated independently according to a  mixture
of two Gaussians \cite{jmlr,LVQwsom,performance}: 
\begin{eqnarray} \label{prob}
 P(\vec{\xi}) &=& {\textstyle \sum_{m=1,2}} \, \, \,p_m  P(\vec{\xi}\mid m)
 \mbox{~~with~contributions} \nonumber \\ 
  P(\vec{\xi}\mid m)  &=& \frac{1}{(2\, \pi\, v_m)^{N/2}}  \, 
 \exp \left[-\frac{1}{2 \, v_m} \left( \vec{\xi} - \lambda 
\vec{B}_m \right)^2 \right].
\end{eqnarray}
The target classification 
coincides with the cluster membership, i.e.\
$\sigma=m$ in Eq.\ (\ref{LVQ1f}).
The class-conditional densities $P(\vec{\xi}\!\mid\!m\!=\!1,2)$ correspond 
to isotro\-pic, spherical Gaussians with variance $\, v_m$ and
mean  $\lambda \, \vec{B}_m$.
Prior weights of the clusters are denoted as 
$p_m$ and satisfy  $p_1 + p_2 =1$.
We assume that the 
vectors $\vec{B}_m$  are orthonormal with  
 $\vec{B}_1^{\, 2}=\vec{B}_2^{\, 2}=1$ and $\vec{B}_1 \cdot \vec{B}_2 =0$.
 Obviously, the classes $m=1,2$ are not perfectly
separable due to the overlap of the clusters. 

We denote conditional averages over $P(\vec{\xi}\mid m)$  by 
$\left\langle \cdots \right\rangle_m $, whereas
mean values  $\langle \cdots 
\rangle = 
\sum_{m=1,2} \, p_m \, \left\langle\cdots\right\rangle_m $
are defined with respect to the full density (\ref{prob}). 
One obtains, for instance,  the conditional and full averages
\begin{eqnarray}  \label{xinorm}
 \left\langle \vec{\xi} \right\rangle_m &=& 
 \lambda \, \vec{B}_m, \mbox{~~~}
 \langle \vec{\xi}^{\, 2} \rangle_m  
   =  v_m \, N  + \lambda^2 \mbox{~and}  \nonumber \\
 \langle \vec{\xi}^{\, 2}\rangle  &=&  
  \left(p_1v_1  +  p_2  v_2 \right) \, N + \lambda^2.   
  \end{eqnarray} 
Note that in the thermodynamic limit $N\to\infty$ 
considered later, $\lambda^2$ can be neglected 
in comparison to the terms of ${\cal{O}}(N)$
in Eq.\ (\ref{xinorm}). 

Similar clustered densities 
have been studied in the context of
unsupervised learning and
supervised perceptron training, see e.g.\ 
\cite{sompo,freking,marangi}.
Also, online LVQ in stationary 
situations was analysed 
in e.g.\ \cite{jmlr}. 

  Here we focus on the question 
  whether LVQ learning schemes are able to cope 
  with drift in characteristic model situations 
  and whether extensions like weight decay
  can improve 
  the performance in such settings.

\subsection{Layered Neural Networks}  \label{SCM} \label{scm} 

The term Soft Committee Machine (SCM) 
has been established for shallow 
feedforward neural networks with 
a single hidden layer and a 
linear output unit, see for instance
\cite{ahr2,gradient,transient,ahr,inoue,backprop,saadsolla1,saadsolla2,catichascm}. 
Its structure resembles that of a \textit{(crisp)}
 committee machine
with binary threshold hidden units, where the 
network output is given by their 
\textit{majority vote},
see \cite{sompo,engel,revmod} and references therein. 

The output of an SCM with $K$ hidden units
and fixed hidden-to-output weights
is  of the form
\begin{equation} \label{scmdef}
y(\vec{\xi}) = 
\sum_{k=1}^K	 \, g(\vec{w}_k \cdot \vec{\xi}) \mbox{~~where~~} \vec{w}_k \in \mathbb{R}^N
\end{equation}denotes the weight
vector connecting the $N$-dimensional input layer 
with the $k$-th hidden unit. A non-linear transfer function
$g(\cdots)$ defines the hidden unit states and the
final output is given as their sum. 

\noindent 
As specific examples we  consider the sigmoidal
\begin{equation}  \label{erfdef}
g(x) = {\rm{erf}}\left(x/\sqrt{2}\right) \mbox{~~with~} g^\prime(x)= \sqrt{{2}/{\pi}} \,\, 
e^{-x^2/2}
\end{equation}
and the popular \textit{rectified linear unit}
(ReLU): 
\begin{equation}  \label{reludef}
g(x) = x \, \Theta(x) \mbox{~~with~} g^\prime(x)=
 \, \Theta(x).
\end{equation}

The activation (\ref{erfdef}) resembles closely other sigmoidal
functions, e.g. the more popular $\tanh(x)$, but it facilitates
the analytical treatment in the mathematical analysis as exploited in 
\cite{gradient}, originally.
In the following we refer to an SCM with the above sigmoidal activation 
as Erf-SCM, for brevity.

Similarly, we use the term ReLU-SCM 
for networks with hidden unit states given by Eq.\ (\ref{reludef}). The ReLU 
activation  has recently gained
significant popularity in the context of Deep Learning \cite{deep}. 
This is, among other reasons, due to its simplicity which offers 
computational ease and numerical stability.
According to the literature, ReLU networks
have displayed favorable training and
generalization behavior in several practical
applications and benchmark problems 
\cite{timetoswish,krizhevsky,acoustic,nair,searching}. 

Note that an SCM,  cf. Eq.\ (\ref{scmdef}), is not quite 
a \textit{universal approximator}. However, this property 
could be achieved by introducing
hidden-to-output weights and 
adaptive local 
thresholds $\vartheta_i \in \mathbb{R}$ in  hidden unit 
activations of the form $g\left(\vec{w}_i\cdot
\vec{\xi} -\vartheta_i\right)$, see 
\cite{cybenko}.  
Adaptive hidden-to-output weights have
been studied in, for instance, \cite{backprop}
from a statistical physics perspective.  
However,  we restrict
ourselves to the simpler model defined above
and focus on basic dynamical effects
and potential differences of ReLU- vs. Erf-SCM in the presence 
of concept drift.

\subsubsection{Regression Scheme and On-Line Learning} 

The training of a neural 
network with real-valued output $y(\vec{\xi})$ based on  examples 
$\left\{ \vec{\xi}^\mu \in \mathbb{R}^N, 
\tau^\mu \in \mathbb{R} \right\}$ 
for a regression problem 
is frequently guided by  the quadratic
deviation of the network output from the target
values \cite{Hastie,Bishop,deep}   . It serves as a cost function
which evaluates the network performance
with respect to a single example as
\begin{equation} \label{scmcost}
 e^\mu \left(\{\vec{w}_k\}_{k=1}^K\right) =
   \frac{1}{2} \big( y^\mu - \tau^\mu \big)^2
   \mbox{~~with~} 
   y^\mu = y(\vec{\xi}^\mu).  
\end{equation}
In stochastic or on-line gradient descent, updates
of the weight vectors are based on the presentation
of  a single example at \textit{time step} $\mu$ 
\begin{equation} \label{genericgradient}
  \vec{w}_k^{\mu} = \vec{w}_k^{\mu-1} + \frac{\eta}{N} \, \Delta\vec{w}_k^{\mu} 
  \mbox{~~with~~~} \Delta\vec{w}_k^\mu =
  \, -  \, 
  \frac{\partial e^\mu}{\partial \vec{w}_k}
\end{equation}
where the gradient is evaluated in $\vec{w}_k^{\mu-1}$.
For the SCM architecture specified in Eq.~(\ref{scmdef}),
\( \partial y^\mu/ {\partial \vec{w}_k} 
  = g'\left(h_k^\mu\right) \vec{\xi}^\mu, \)
and we obtain
  \begin{equation} \label{scmgradient2}
  \Delta\vec{w}_k^{\mu} =  - 
  \left(\sum_{i=1}^K
  g\left(h_i^\mu\right) - \tau^\mu \right)  \, 
  g^\prime \left(h_k^\mu\right) \vec{\xi}^\mu 
\end{equation}
with the inner products $h^\mu_i =
\vec{w}_i^{\mu-1}\cdot\vec{\xi}^\mu$ 
of the current weight vectors
with the next example input in the stream. Note that the change
of weight vectors is proportional to $\vec{\xi}^\mu$ 
and can be interpreted as a form of \emph{Hebbian 
Learning} \cite{Bishop,deep,Hastie}.  

\subsubsection{Student-Teacher Scenario and Model Data}
In order to define and model meaningful learning 
situations we resort to the consideration of 
student-teacher scenarios \cite{sompo,handbook,engel,revmod}.

We assume 
that the target  can be defined in terms of
an SCM with a number $M$ of hidden units
and a specific set of weights 
$\left\{\vec{B}_m \in \mathbb{R}^N \right\}_{m=1}^M$: 
\begin{equation} \label{scmteacherdef}
\tau(\vec{\xi}) = 
\sum_{m=1}^M	 \, g(\vec{B}_m \cdot \vec{\xi})
\mbox{~~and~} 
\tau^\mu = \tau(\vec{\xi}^\mu) = \sum_{m=1}^M g(b_m^\mu) 
\end{equation}
with 
 $b_m^\mu = \vec{B}_m \cdot \vec{\xi}^\mu$ 
for one of the training examples. 
This so-called teacher network can be equipped
with $M>K$ hidden units in order to model regression
schemes which cannot be learnt by an SCM student
of the form (\ref{scmdef}). On the contrary,
$K>M$ would correspond to an {\sl over-learnable}
target or {\sl over-sophisticated} student. For the discussion of these
highly interesting cases in stationary environments, see
for instance 
\cite{gradient,transient,backprop,saadsolla1,saadsolla2}.
In a student-teacher scenario with $K$ and
$M$ hidden units
the update of the student 
weight vectors by on-line gradient descent
is given by Eq. (\ref{scmgradient2}) with 
$\tau^\mu$ from Eq.\ (\ref{scmteacherdef}).

In the following, we will restrict our analysis to 
perfectly matching student
complexity with
$K=M=2$ only, which further simplifies Eq. (\ref{scmgradient2}). Extensions to more hidden units and settings with
$K\neq M$ will be considered in forthcoming projects.

In contrast to the model for LVQ-based
 classification, the  vectors $\vec{B}_m$ define the 
 target outputs $\tau^\mu= \tau(\vec{\xi}^\mu)$ explicitly
via the teacher network for any input vector. 
While clustered input densities of the form (\ref{prob}) 
can also be studied for feedforward networks as in \cite{marangi,meir},
we assume here that the actual input vectors 
are uncorrelated with the teacher vectors 
$\vec{B}_m$.  Consequently, we can resort to a  
simpler model density and 
consider vectors $\vec{\xi}$ of independent,
zero mean, unit variance components with 
\begin{equation} \label{spherical}
P(\vec{\xi}) = {(2\, \pi)^{-N/2}}  \, 
 \exp \left[- \, \vec{\xi}^2/2 \right].
\end{equation}
Note that the density (\ref{spherical}) is recovered formally from 
Eq.\ (\ref{prob}) by setting  $\lambda=0$ and $v_1=v_2=1$,
for which both clusters in (\ref{prob}) coincide in the
origin and the parameters $p_{1,2}$ become irrelevant.

{Note that the student/teacher scenario considered
here is different from concepts used in studies of 
\textit{knowledge distillation}, see \cite{distillation} 
and references therein. In the context of distillation,
a teacher network is itself trained on a given data set to approximate the target function. Thereafter a student
network, frequently of a simpler architecture, 
distills the knowledge in a subsequent training process.  
In our work, as in most statistical physics
based studies \cite{sompo,engel,revmod}, the teacher 
network is taken to 
directly define the true target function. 
A particular architecture is
chosen and, together with its fixed weights, it controls the complexity of the task. 
The teacher network provides correct target outputs 
to all input data that are generated according to the
distribution in Eq.~(\ref{spherical}). In the
actual training process, a sequence of such
input vectors and teacher-generated labels is presented to the student network. } 

\subsection{Mathematical analysis of the training dynamics} \label{mathdynamics} 

\label{dynamics} \label{math}    

In the following we sketch the successful theory of 
on-line learning \cite{sompo,handbook,engel,cambridge,revmod}
as, for instance, applied to the dynamics of LVQ algorithms in 
\cite{jmlr,performance,LVQwsom}
and to on-line gradient descent in SCM in 
\cite{gradient,transient,inoue,backprop,saadsolla1,saadsolla2,catichascm}. 
We refer the
reader to the original publications for details. 
The extensions to non-stationary situations with
concept drifts are discussed in Sec.\ 
\ref{nonstat}.

The mathematical analysis proceeds along the same generic steps
in both settings. Our presentation follows closely 
the descriptions in \cite{wsomdrift} and \cite{entropy}. 

We consider adaptive vectors
$\vec{w}_{1,2}\in \mathbb{R}^N$ (prototypes in LVQ,
student weights in the SCM) while the 
characteristic vectors $\vec{B}_{1,2}$ 
specify the target task (cluster centers in LVQ
training, SCM teacher vectors for regression). 

The consideration of the {\it thermodynamic limit} 
$N\to\infty$ is 
instrumental for the theoretical treatment.
The limit facilitates 
the following key steps which, eventually, yield an exact mathematical description
of the training dynamics in terms of ordinary differential equations (ODE):

\noindent 
(a) \textit{Order parameters} \ \\
The many degrees of freedom, i.e. the components of
the adaptive vectors,  can be
  characterized  in terms of only very few quantities.
  The definition of these so-called {\it order parameters} follows naturally from the
   mathematical structure of the model.
After presentation of a number $\mu$ of examples,
as indicated by corresponding superscripts,  we 
describe
the system by the projections for
$ 
 i,k,m \in \{1,2\}
$ 
 \begin{equation} \label{orderdef}
  R_{im}^\mu=\vec{w}_i^\mu \cdot \vec{B}_m
 \,\, \mbox{and~~}
  Q_{ik}^\mu=\vec{w}_i^\mu \cdot \vec{w}_k^\mu.
 \end{equation} 
 Obviously,  $Q_{11}^\mu,Q_{22}^\mu$ and 
 $Q_{12}^\mu=Q_{21}^\mu$ relate to the norms and mutual overlap of 
 the adaptive vectors, while 
 the quantities $R_{im}$ specify their projections
 into the linear subspace defined by the
 characteristic vectors
 $\{\vec{B}_1,\vec{B}_2\}$, respectively. 
 
 \noindent 
 (b) \textit{Recursions} \ \\
 Recursion relations for the order parameters (\ref{orderdef}) can
 be derived directly
 from the update steps, which are of the generic form~
 $
 \vec{w}_k^\mu \, = \vec{w}_k^{\mu-1} \, + 
  \eta/N \, \Delta \vec{w}_k^\mu. $
 The corresponding inner products yield
 \begin{eqnarray} \label{recursions}
  N({R_{im}^{\mu} - R_{im}^{\mu-1}})
  & = & \eta  \, \Delta\vec{w}_i^\mu \cdot\vec{B}_m   \nonumber \ \\
  N ({Q_{ik}^{\mu} - Q_{ik}^{\mu-1}})
  & = & \eta\left( \vec{w}^{\mu-1}_i
   \cdot \Delta \vec{w}^{\mu}_k 
   + \vec{w}^{\mu-1}_k \cdot \Delta \vec{w}^{\mu}_i
   \right) \nonumber \\
   &+&  \, \eta^2/N \, \Delta\vec{w}^{\mu}_i
     \cdot \Delta\vec{w}^{\mu}_k.
   \end{eqnarray}   
 Terms of order ${\cal O}(1/N)$ on the r.h.s.\ 
 will be neglected in the following. Note
 however that  
 $\Delta\vec{w}^{\mu}_i
     \cdot \Delta\vec{w}^{\mu}_k$ 
     comprises contributions of order 
     $ |\vec{\xi}|^2 \propto N$ for the considered updates
     (\ref{generic}) and (\ref{genericgradient}).
     
\noindent    
(c)  \textit{Averages over the Model Data} \ \\
 Applying the central limit theorem  (CLT)
 we can perform an average over the random sequence of  
 independent examples.
 
 Note that  $\Delta\vec{w}^\mu_k \propto \vec{\xi}^\mu$ 
 or $\Delta\vec{w}^\mu_k \propto \left(\vec{\xi}^\mu -
 \vec{w}^{\mu-1}_k\right)$ for the SCM and LVQ, respectively.    
 Consequently,  the current  input $\vec{\xi}^\mu$ enters the r.h.s.\ 
 of Eq.\ (\ref{recursions}) only through its 
 norm $\mid \vec{\xi}\mid^2 = {\cal{O}}(N)$ and the quantities 
 
 \begin{equation} \label{hb}
  h_i^\mu \, = \vec{w}_i^{\mu-1} \cdot \vec{\xi}^\mu  \mbox{~~and~~}
  b_m^\mu \, = \vec{B}_m \cdot \vec{\xi}^\mu.
 \end{equation}
Since these inner products correspond to sums of many 
independent random quantities in our model, the CLT implies
 that the projections in Eq.\ (\ref{hb})  are correlated Gaussian 
 quantities for large $N$ and 
 the joint density $P(h_1^\mu,h_2^\mu,b_1^\mu,b_2^\mu)$  is given 
 completely  by first and second moments.

{\it LVQ:~} 
For the clustered density, cf.
 Eqs. (\ref{prob}), the conditional moments read
 \[
 \left\langle h^\mu_{i} \right\rangle_{m} = \lambda
 R_{im}^{\mu-1},
 \quad
  \left\langle b^\mu_{m} \right\rangle_{n} = \lambda
  \delta_{mn},
 \]
 \[
  \left\langle h^\mu_{i} h^\mu_{k} \right\rangle_{m} -
  \left\langle h^\mu_{i} \right\rangle_{m}
  \left\langle  h^\mu_{k} \right\rangle_{m} = 
  v_m \, Q^{\mu-1}_{ik},
 \]
 \[
  \left\langle h^\mu_{i}  b^\mu_{n} \right\rangle_{m} -
  \left\langle h^\mu_{i} \right\rangle_{m}
  \left\langle  b^\mu_{n} \right\rangle_{m} = 
  v_m \, R^{\mu-1}_{in},
 \]
 \begin{equation} \label{moments}
  \left\langle b^\mu_{l} b^\mu_{n} \right\rangle_{m} -
  \left\langle b^\mu_{l} \right\rangle_{m}
  \left\langle  b^\mu_{n} \right\rangle_{m} = v_m \, \delta_{ln},
  \end{equation}
 \noindent  with $i,k,l,m,n \in \{1,2\}$ and 
 the Kronecker-Delta $\delta_{ij}= 1$ for $i=j$ and $\delta_{ij}=0$ else. 

 {\it SCM:~}
 In the simpler case of the isotropic, spherical
 density (\ref{spherical}) with $\lambda=0$ and
 $v_1=v_2=1$ the moments reduce to
 \begin{equation} \label{isotropicmoments} 
 \left\langle h^\mu_{i} \right\rangle = 0,
 \,
  \left\langle b^\mu_{m} \right\rangle = 0,
  \left\langle h^\mu_{i} h^\mu_{k} \right\rangle -
  \left\langle h^\mu_{i} \right\rangle
  \left\langle  h^\mu_{k} \right\rangle = 
    Q^{\mu-1}_{ik}
\end{equation} 
\[
  \left\langle h^\mu_{i}  b^\mu_{n} \right\rangle -
  \left\langle h^\mu_{i} \right\rangle
  \left\langle  b^\mu_{n} \right\rangle = 
 R^{\mu-1}_{in},
  \left\langle b^\mu_{l} b^\mu_{n} \right\rangle \!-\!
  \left\langle b^\mu_{l} \right\rangle
  \left\langle  b^\mu_{n} \right\rangle = \delta_{ln}.
\]
Hence, in both cases (LVQ and SCM)  the four-dim.\ density of $h_{1,2}^\mu$ 
  and $b_{1,2}^\mu$ is fully specified
  by the values of the order parameters in the 
  previous time step and the parameters of the model density. 
  This important
  result enables us to average the recursion relations
  (\ref{recursions}) over the most recent training example by means 
  of Gaussian integrals. The resulting r.h.s.\
  can be expressed as functions of 
   $\{ R_{im}^{\mu-1},Q_{ik}^{\mu-1} \}.$ 
  Obviously, the precise form depends
  on the details of the algorithm and model setup. \ \\

\noindent (d) \textit{Self-Averaging Properties}  \ \\ 
 The self-averaging property of the
 order parameters allows us to describe the 
 dynamics in terms of mean values: Fluctuations
 of the stochastic dynamics can be neglected in the
 limit $N\to\infty$. The concept relates to  
 the statistical physics of disordered materials and
 has been transferred successfully to the 
 study of neural network models and learning 
 processes \cite{sompo,engel,revmod}. A detailed
 mathematical discussion in the context of sequential 
 on-line learning dynamics is given in \cite{reents}.
 As a consequence, we can interpret the averaged equations
 (\ref{recursions}) directly as deterministic recursions 
 for the actual values of $\{R_{im}^\mu,Q_{ik}^\mu \},$ which
  coincide with their disorder average in the thermodynamic
  limit. \\
  
\noindent (e) \textit{Continuous Time Limit} \ \\
 In the thermodynamic limit $N\to\infty,$
 ratios of the form $(\ldots)/(1/N)$ 
 on the left hand sides of Eq.\ (\ref{recursions}) can
  be interpreted as derivatives
  with respect to a continuous learning time $\alpha$
  defined by
  \begin{equation} \label{alphadef}
   \alpha \, = {\, \mu\, }/{N} \mbox{~with~} 
   d\alpha \, \sim \, 1/N. 
  \end{equation} 
  This scaling corresponds to the natural assumption that  
  the number of examples should be 
   proportional to the number of adaptive quantities in
   the system. 
  
 Averages are performed over the joint density
 $P\left( h_1^\mu,h_2^\mu,b_1^\mu,b_2^\mu \right)$ 
 corresponding to the latest, independently 
 drawn input vector. For simplicity, we
 omit indices $\mu$ in the following.
 The resulting sets of coupled ODE is of the form
  \begin{equation} \label{odegeneric} 
   \left[\frac{dR_{im}}{d\alpha} \right]_{stat} \!\!\!\!\!
    = 
    \eta F_{im} 
  \mbox{~;~} 
  \left[\frac{dQ_{ik}}{d\alpha}\right]_{stat} \!\!\!\!\!
    = 
    \eta \,  G^{(1)}_{ik}  + \eta^2 G^{(2)}_{ik}.
  \end{equation} 
 Here, the subscript {\it stat} indicates that the ODE describe
 learning from a stationary density, Eqs.  (\ref{prob}) or (\ref{spherical}).  
 
 \textit{Limit of small learning rates:} \\
 The dynamics can also be studied in the limit of small learning
 rates $\eta\to 0$. In this case, the term  $\eta^2 G_{ik}^{(2)}$
 can be neglected in Eq.\ (\ref{odegeneric}). 
 In order to retain non-trivial performance, the small
 step size has to be compensated for by training with a large number of examples 
 that diverges like $1/\eta$. Formally, we introduce the  quantity 
 $\widetilde{\alpha}$ in the simultaneous limit 
 \begin{equation} \label{alphatilde} 
 \widetilde{\alpha} \, = \lim_{\eta\to 0} \lim_{\alpha\to \infty}  \, (\eta \alpha),
 \end{equation} 
 which leads to a simplified system of ODE 
 \begin{equation} \label{odesmallrate} 
   \left[\frac{dR_{im}}{d\widetilde{\alpha}} \right]_{stat} \!\!\!\!\!
    = 
     F_{im} 
  \mbox{~;~} 
  \left[\frac{dQ_{ik}}{d\widetilde{\alpha}}\right]_{stat} \!\!\!\!\!
    = 
      G^{(1)}_{ik}
  \end{equation} 
 in rescaled continuous time $\widetilde{\alpha}$ for $\eta\to 0.$

\textit{LVQ:} In the classification model we have to  insert
  \[
   F_{im}
    =  \left(\left\langle b_m  f_i \right\rangle \! -\!  
    R_{im}  \left\langle f_i \right\rangle \right), \,
 \]
  \[
   G^{(1)}_{ik}
    = \Big(
     \left\langle h_i f_k  + h_k  f_i \right\rangle
    \! -\!  Q_{ik}  \left\langle f_i \! +\! f_k \right\rangle \Big) 
    \]
    \begin{equation}   \label{FGLVQ}
   \mbox{and~} G^{(2)}_{ik}=  {\textstyle \sum_{m=1,2}} \,  v_m p_m
   \left\langle f_i  f_k \right\rangle_m
  \end{equation}
  in Eqs.\ (\ref{odegeneric}) or (\ref{odesmallrate}). 
  The LVQ1 modulation functions $f_i$ is given in 
  Eq.\ (\ref{LVQ1f}) and conditional averages $\langle\ldots\rangle_m$
  are with respect to the density (\ref{prob}).

   \textit{SCM:} In the case of non-linear regression  we obtain
   \[
   F_{im}
    = \langle \rho_i b_m \rangle, \quad 
   G^{(1)}_{ik}
    = \langle \left(\rho_{i} h_k  + \rho_k  h_i\right)\rangle,
    \]
    \begin{equation}  \label{FGSCM} 
     \mbox{and~} 
   G^{(2)}_{ik}= \langle \rho_i \rho_k \rangle  
   \mbox{~with~} \rho_k=-(y-\tau) g^\prime(h_k).
  \end{equation}
 Eventually, 
 the r.h.s.\ of Eqs.\ (\ref{odegeneric}) or (\ref{odesmallrate}) are expressed
 in terms of elementary functions of order parameters. 
 For the straightforward, yet lengthy results
 we refer the  reader to the original
 literature for LVQ  \cite{jmlr,performance} 
 and SCM \cite{transient,backprop,saadsolla1,saadsolla2}, 
 respectively.  \ \\

\noindent (f) \textit{Generalization error} \ \\ 
   After training, the success of learning is
  quantified  in terms of the generalization  error
  $\epsilon_g$, which is also given as
  a function of the macroscopic order parameters. 

  \textit{LVQ:~} 
  In the case of the LVQ model, $\epsilon_g$
   is given as
  the probability of misclassifying a novel, randomly drawn
  input vector. The class-specific errors corresponding
to data from clusters
$k=1,2$ in Eq.\ (\ref{prob}) can be considered separately:
\begin{equation} \label{eg} 
 \epsilon_g  =   p_1 \, \epsilon_g^1  +  p_2 \, \epsilon_g^2 
  \mbox{~where~}  \epsilon_g^k \, = \,   \bigg\langle 
     \Theta\left( d_{k} - d_{\widehat{k}} \right) 
   \bigg\rangle_k
\end{equation}
is the class-specific  misclassification rate, i.e.\ the  probability 
for an example drawn from a cluster $k$ to be assigned
to $\widehat{k}\neq k$ with $d_{k} > d_{\widehat{k}}$. 
For the derivation of  the 
class-wise and total generalization error
for systems with two prototypes
as functions of the order parameters 
we also refer to \cite{jmlr}. One obtains
\begin{equation}
\epsilon_g^k \, = \, \Phi \left( 
\frac{
Q_{kk}-Q_{\widehat{k}\widehat{k}}-2\lambda 
( R_{kk}-R_{\widehat{k}\widehat{k}})}  {2 \sqrt{v_k} 
\sqrt{Q_{11}-2Q_{12}+ Q_{22}}} \right) 
\end{equation} 
with the function $\Phi(z)=\int_{-\infty}^{z}
dx \, {e^{-x^2/2}}/{\sqrt{2\pi}}.$ 
 
\textit{SCM:~}
In the regression scenario, the generalization error is
defined as an average $\left\langle \cdots \right\rangle$ of the quadratic deviation 
between student and teacher output over
the isotropic density, cf.\ Eq.\ (\ref{spherical}): 
  \begin{equation} \label{egscm} 
 \epsilon_g  \, =  \frac{1}{2}
   \left\langle 
     \left[\sum_{k=1}^K g
     \left({h_k}\right)
     - \sum_{m=1}^M g\left({b_m}\right) \right]^2
   \right\rangle.
\end{equation}

\noindent 
In the simplifying case of 
$K=M=2$ we obtain for \\ 
Erf-SCM: 
 \[
 \epsilon_g  \, =  \frac{1}{3} + \frac{1}{\pi}
 \ \sum_{i,k=1}^2 \sin^{-1}\left(\frac{Q_{ik}}
 {\sqrt{1+Q_{ii}}\sqrt{1+Q_{kk}}}\right)
 \]
 \begin{equation}
 \mbox{~~~~~~~~} \label{egscm2_sigmoidal} 
 - \frac{2}{\pi} \sum_{i,m=1}^2 \sin^{-1}\left(\frac{R_{im}}{\sqrt{2} 
 \sqrt{1+Q_{ii}} } \right) \end{equation} 
and for  ReLU-SCM:
\[
 \label{egscm2_relu}
 \epsilon_g\!\!=\!\!\sum_{i,j=1}^2 \!\!\left[
 \frac{Q_{ij}}{8}\!+\!\frac{\sqrt{Q_{ii}Q_{jj}\!-\!Q_{ij}^2}\!+\! 
 Q_{ij}\sin^{-1}\left(\!\frac{Q_{ij}}{\sqrt{Q_{ii}Q_{jj}}}\!\right)}{4\pi}  \right]
 \]
 \begin{equation}-\!\!\sum_{i,j=1}^2 \!\!\left[ \frac{R_{ij}}{4}\!\!+\!\!\frac{\sqrt{Q_{ii}\!-\!R_{ij}^2}\!+\! 
 R_{ij}\sin^{-1}\left(\frac{R_{ij}}{\sqrt{Q_{ii}}}\!\right)}{2\pi} \right]
 \!+\! \frac{\pi\!+\!1}{2\pi}.
 \end{equation} 
Both results are for orthonormal teacher vectors, extensions to 
general  $\vec{B}_m \cdot \vec{B}_n = T_{mn}$ can be found in 
\cite{saadsolla2,entropy}. 

 
 \noindent 
 (g) \textit{Learning curves} \ \\
The (numerical) integration of the ODE for a given
 particular training algorithm, model density and
 specific initial conditions
  $\{ R_{im}(0), Q_{ik}(0) \}$ yields the temporal
  evolution of order parameters 
  in the course of training. 
  
Exploiting the self-averaging properties of  order parameters  once more, 
we can obtain the learning curves 
$\epsilon_g (\alpha)=
\epsilon_g\left(\{ R_{im}(\alpha), Q_{ik}(\alpha)\}\right)$ or 
the class-wise  $\epsilon_g^{k}(\alpha)$, respectively. 
Hence, we  determine the typical generalization error after 
on-line training with $(\alpha \, N)$ random examples.


\subsection{The Learning Dynamics Under Concept Drift}  \label{nonstat} 

The analysis summarized in the previous section concerns 
learning in the presence of a stationary concept, 
i.e.\ for a density of
the form (\ref{prob}) or (\ref{spherical}) 
which does not change in the course of  
training. 
Here, we introduce the effect of concept 
drift to the modelling framework and consider weight decay as an 
example mechanism for explicit \emph{forgetting}.


\subsubsection{Virtual Drift in Classification}
As defined above, virtual drifts 
affect statistical properties of the observed example data
while  the actual target function remains unchanged.

A variety of virtual drift processes can be addressed 
in our modelling framework. 
For example, time-varying \textit{label noise}
in regression or classification 
could be incorporated in a straightforward
way \cite{sompo,engel,revmod}. Similarly, 
non-stationary cluster variances 
in the input density, cf. Eq.  (\ref{prob}), can be 
introduced through explicitly time-dependent $v_\sigma(\alpha)$ 
into Eq. (\ref{odegeneric}) for the LVQ system. 

Here we focus on  
a particularly relevant case in classification,
in which a varying fraction of examples represents each of the 
classes in the data stream. We consider 
non-stationary, $\alpha$-dependent prior probabilities 
$p_1(\alpha) = 1-p_2(\alpha)$ in the mixture density 
(\ref{prob}). 
In practical situations, varying class bias can complicate 
the training significantly and lead to inferior performance
\cite{Wang}. 
Specifically, we distinguish the following scenarios: 
 \paragraph{(A) Drift in the training data only} \ \\
 Here we assume 
 that the true target classification
 is defined by a fixed 
 \textit{reference density}
 of data. 
 As a  simple example we consider
 equal priors $p_1=p_2=1/2$ in a  
 symmetric reference density
 (\ref{prob}) with $v_1=v_2$.  
 On the contrary, the  characteristics 
 of the observed 
 training data are assumed to be
 time-dependent. In particular, we study 
 the effect of non-stationary
 $p_m(\alpha)$ and weight decay on the
 learning dynamics.
 Given the order parameters of the learning 
 systems in the course of training,
 the corresponding \textit{reference
 generalization error}
 \begin{equation} \label{epsref}
 \epsilon_{ref}(\alpha)=
 \left(\epsilon_g^1 +  \epsilon_g^2\right)/2
 \end{equation} 
 is obtained
 by setting $p_1=p_2=1/2$
 in Eq. (\ref{eg}), but inserting 
 $R_{im}(\alpha)$ and $Q_{ik}(\alpha)$
 as obtained from the integration of the 
 corresponding ODE with time dependent
 $p_1(\alpha)=1-p_2(\alpha)$ in the training 
 process.

 \paragraph{(B) Drift in training and test data} \ \\
 In the second interpretation we assume that the  variation of $p_m(\alpha)$ 
 affects training and  test data in the same way.  
 Hence, the change of the  statistical properties of
 the data is inevitably accompanied  by a modification of  the target 
 classification: For instance, the  Bayes optimal classifier and
 its best linear approximation depend explicitly on the actual priors 
 \cite{jmlr}.

 The learning system is supposed to 
 track the actual drifting concept and we refer to 
 the corresponding generalization error as the
 \textit{tracking error} 
 \begin{equation} \label{epstrack} \epsilon_{track}=
 p_1(\alpha) \, \epsilon_g^1 \,
 +\, p_2(\alpha) \, \epsilon_g^2.
 \end{equation}  
 
 In terms of modelling the training dynamics, 
 both scenarios, (A) and (B), require the 
 same straightforward
 modification of the ODE system: the
 explicit introduction of $\alpha$-dependent
  quantities $p_\sigma(\alpha)$ 
 in Eq.\ (\ref{odegeneric}). The obtained
 temporal evolution yields the 
 reference error $\epsilon_{ref}(\alpha)$ for
 the case of drift in the training data (A) and 
 $\epsilon_{track}(\alpha)$ in 
 interpretation (B). 

Note that in both interpretations, we consider the very same
drift processes affecting the training data. However, the interpretation
of the relevant performance measure is different. In (A) only the
training data is subject to the drift, but the classifier is 
evaluated with respect to an idealized static situation representing a fixed
target. On the contrary, 
the tracking error in (B) is thought to be computed with respect to 
test data available from the stream, at the given time. 
Alternatively, one could interpret (B) as an example of
real drift with a non-stationary target, where $\epsilon_{track}$ represents
the corresponding generalization error. However, 
we will refer to (A) and (B) as virtual drift throughout the following. 

\subsubsection{Real Drift in Regression} \label{realdrift}
In the presented framework, a real drift can be modelled
as a process which displaces the characteristic
vectors $\vec{B}_{1,2}$, i.e.\ the cluster centers in LVQ
or the 
teacher weight vectors in the SCM.  Here we focus on the latter case
and refer the reader to \cite{entropy} for earlier results on 
LVQ training under real drift. 

A variety of time-dependences 
could be considered in the model. We 
restrict ourselves to the
analysis of diffusion-like random displacements  of 
 vectors $\vec{B}_{1,2} (\mu)$ at each time step. 
 Upon presentation of example $\mu$, we assume that random
 vectors $\vec{B}_{1,2}(\mu)$ are generated which satisfy
 the conditions
 \[
 \vec{B}_1(\mu) \cdot \vec{B}_1(\mu\!-\!1)  =  
 \vec{B}_2(\mu) \cdot \vec{B}_2(\mu\!-\!1)  = \left(1 - {\delta}/{N}\right) 
 \]
 \begin{equation}\label{Bdrift}
 \vec{B}_1(\mu)\cdot\vec{B}_2(\mu)= 0 \mbox{~and~} \mid \vec{B}_1(\mu)\mid^2 =  \mid \vec{B}_2(\mu)\mid^2 = 1. 
   \end{equation}
 Here $\delta$ quantifies the strength of the drift process. The displacement
 of the teacher vectors is very small in an individual training step.
 For simplicity we assume that the orthonormality of teacher vectors
 is preserved in the drift. 
 In continuous 
 time $\alpha=\mu/N$,
 the drift parameter defines a characterstic scale $1/\delta$  on which 
 the overlap of the current teacher vectors with their 
 initial positions 
 decay: 
 $ \label{memory} 	
 \vec{B}_{m}(\mu)\cdot \vec{B}_{m}(0)\, =
  \exp[-\delta \, \mu/N ].
 $

The effect of such a drift process is easily taken into 
account in the formalism: 
For a particular student $\vec{w}_i\in \mathbb{R}^N$ we obtain \cite{drifting1,drifting2,caticha,caticha2} 
\begin{equation}	
 \left[\vec{w}_i\cdot\vec{B}_k(\mu)\right]  =
 \left(1- {\delta}/{N}\right) \, 
 \left[\vec{w}_i\cdot\vec{B}_k(\mu-1)\right]. 
\end{equation}
under the above specified random displacement. 
Hence, the drift
tends to decrease the quantities $R_{ik}$
which clearly reduces the 
success of training compared with the case of
stationary teachers. 
The corresponding ODE in the limit
$N\to\infty$ 
in the drift process (\ref{Bdrift}) become
 \[
   \left[{dR_{im}}/{d\alpha} \right]_{drift} \, = \, 
   \left[{dR_{im}}/{d\alpha} \right]_{stat}  \, - \delta \, R_{im}
   \mbox{~~and}
   \]
  \begin{equation}  \label{odedrift}  
   \left[{dQ_{ik}}/{d\alpha}\right]_{drift} =
  \left[{dQ_{ik}}/{d\alpha}\right]_{stat} 
  \end{equation}
 with the terms $\left[\cdots\right]_{stat}$ for stationary 
 environments taken from Eq.\ (\ref{odegeneric}). 
 Note that now order
 parameters $R_{im}(\alpha)$ correspond to the inner products 
 $\vec{w}_i^\mu\cdot\vec{B}_m(\alpha)$, 
 as the teacher vectors themselves are time-dependent.

\subsubsection{Weight Decay} \label{weightdecay}
 Possible motivations for the introduction of so-called 
 {\it weight decay}  in machine learning systems range
 from  {\it regularization} as 
 to reduce the risk of \emph{over-fitting} in regression and classification 
 \cite{Hastie,Bishop,deep} 
 to the modelling of 
 {\it forgetful memories} in attractor neural networks \cite{mezard,hemmen}.

Here we include weight decay as to enforce 
 \textit{explicit  forgetting} and to potentially 
improve the performance of the systems in the presence
of real concept drift. We consider the multiplication 
of all adaptive vectors by a factor $(1-\gamma/N)$ before the
generic learning step given by $\Delta \vec{w}_i^\mu$  in Eq.\ (\ref{generic}) or Eq. (\ref{genericgradient}) is performed:  
\begin{equation}  \label{withdecay}
  \vec{w}_i^\mu \, 
    = \,  \left(1-{\gamma}/{N}\right) \,  \vec{w}_i^{\mu-1} \, +  {\eta}/{N} \, \Delta \vec{w}_i^\mu.
 \end{equation}

 Since the multiplications with $\left(1-\gamma/N\right)$ 
 accumulate in the course of training, weight decay enforces an increased influence of the most recent training data
 as compared to {\it earlier} examples.  Note that analagous modifications of 
 perceptron training under concept drift have been discussed
 in \cite{drifting1,drifting2,caticha,caticha2}. 
 
 In the thermodynamic limit $N\to\infty$, 
 the modified ODE for training under 
 real drift, cf.\  Eq. (\ref{Bdrift}),  and weight decay, Eq. (\ref{withdecay}), 
 are obtained as 
 \[  
   \left[{dR_{im}}/{d\alpha} \right]_{decay} =
   \left[{dR_{im}}/{d\alpha} \right]_{stat}  - (\delta+\gamma) R_{im}
  \mbox{~~~and} 
  \]
  \begin{equation}  \label{odedecay}  
   \left[{dQ_{ik}}/{d\alpha}\right]_{decay} \,  =
  \left[{dQ_{ik}}/{d\alpha}\right]_{stat}  - 2\, \gamma \,Q_{ik} 
  \end{equation}
 where the terms  for stationary environments in 
 absence of weight decay are given in Eq.\ (\ref{odegeneric}).

\section{Results and Discussion} \label{sec:results}

Here we present and discuss  our results obtained by 
integrating the systems of ODE 
with and without weight decay under 
different time-dependent drifts. For comparison,
averaged learning curves obtained by means
of Monte Carlo simulations are also shown. {These simulations
of the actual training process 
provide an independent confirmation of the ODE-based description and 
demonstrate the relevance of results obtained in the thermodynamic
limit $N\to\infty$ for relatively small, finite systems. }

\begin{figure}[t]
\includegraphics[width=0.48\textwidth]{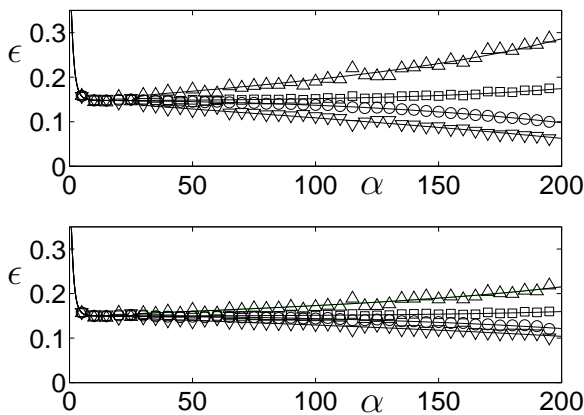}
\put(-98,90){\Large $\alpha$}
\put(-98,8){\Large $\alpha$}
\put(-230,140){\Large $\epsilon$}
\put(-230,56){\Large $\epsilon$}
\caption{ \label{linplot} 
LVQ1 in the presence of a concept drift with 
linearly increasing $p_1(\alpha)$ given by  $\alpha_o\!=\!20$,
$\alpha_{end}\!=\!200$, $p_{max}\!=\!0.8$ in (\ref{linincrease}). 
Solid lines correspond to the integration of ODE with
initialization as in Eq.\  (\ref{initialconditions}). 
We set $v_{1,2}\!=\!0.4$ and $\lambda=1$ in the density (\ref{prob}). 
The upper graph corresponds to LVQ1 without 
weight decay, the lower graph displays results for
 $\gamma=0.05$  in Eq.\ (\ref{withdecay}). In addition, 
Monte Carlo results for $N=100$ are shown: 
class-wise errors $\epsilon^{1,2}(\alpha)$  
are displayed as downward (upward) triangles, 
respectively;  squares mark 
 the reference error $\epsilon_{ref}(\alpha);$ circles correspond to 
 $\epsilon_{track}(\alpha)$, cf.\ Eqs.\
 (\ref{epsref},\ref{epstrack}). }
\end{figure}

\subsection{Virtual Drift in LVQ training} 

{All results presented in the following are for constant
learning rate $\eta=1$ in the LVQ training. The results
remain qualitatively the same for a range of  learning rates.} LVQ
 prototypes were
initialized as normalized independent random vectors
without prior knowledge: 
\begin{equation}  
 Q_{11}(0)=Q_{22}(0)=1, \, 
Q_{12}(0)=0,  \mbox{~and~}
\label{initialconditions}
R_{ik}(0)=0.
\end{equation} 
We study three specific scenarios for the time-dependence $p_1(\alpha)=1\!-p_2(\alpha)$
as detailed in the following.

\subsubsection{Linear increase of the bias} 
Here we consider a time-dependent bias of the form
$ p_1(\alpha) = 	 
     1/2  \mbox{~for~~} \alpha<\alpha_o$ and  
     \begin{equation} \label{linincrease} 
     p_1(\alpha) =
     \frac{1}{2} + \frac{(p_{max}\!-\!1/2) \, (\alpha-\alpha_o)}{(\alpha_{end}-\alpha_o)}  
     \mbox{~for~~} \alpha\geq\alpha_o.
 \end{equation} 
 where the maximum class weight $p_1=p_{max}$ 
 is reached at learning time $\alpha_{end}$. 
Fig.\ \ref{linplot}  shows the 
learning curves as obtained by numerical
integration of the ODE together with Monte Carlo
simulation results for $(N=100)$-dimensional
inputs and prototype vectors. 
As an example we set the parameters to
$ \alpha_o=25, p_{max}=0.8, \alpha_{end}=200$.
The learning curves are displayed for 
LVQ1 without weight decay (upper) and with $\gamma=0.05$ (lower panel).
Simulations show excellent
agreement with the ODE results.

The system adapts to the increasing imbalance
of the training data, as reflected by a
decrease (increase) of the class-wise error for 
the
over-represented  (under-represented) class,
respectively. 
The weighted over-all error
 $\epsilon_{track}$ also decreases, i.e. the 
 presence of class bias facilitates smaller
 total generalization error, see \cite{jmlr}. 
 The performance with respect to unbiased
 reference data deteriorates slightly, i.e.
 $\epsilon_{ref}$ grows with increasing class bias
 as the training data represents the target
 less faithfully.

\subsubsection{Sudden change of the class bias}

\begin{figure}[t]
\includegraphics[width=0.48\textwidth]{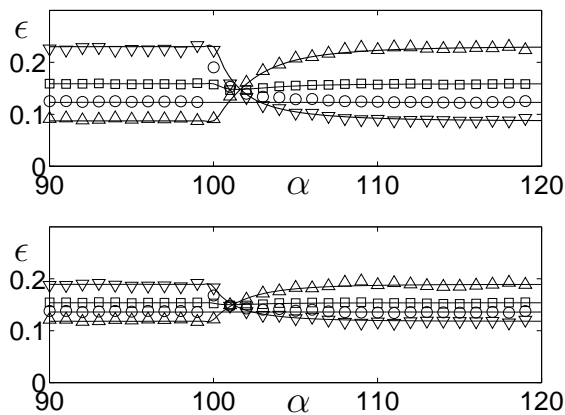}
\put(-118,90){\Large $\alpha$}
\put(-118,8){\Large $\alpha$}
\put(-220,149){\Large $\epsilon$}
\put(-220,65){\Large $\epsilon$}
\caption{ \label{suddenplot} 
LVQ1 in the presence of a concept drift 
with a sudden change of class 
weights according
to Eq.\ (\ref{sudden}) with 
$\alpha_o=100$  and $p_{max}=0.75$. 
Only the $\alpha$-range close
to $\alpha_o$ is shown.
All other details are provided in Fig.\ \ref{linplot}.} 
\end{figure}

Here we consider an instantaneous switch
from   low bias $p_1(\alpha)= 1-p_{max}$
for $\alpha\leq \alpha_o$ to high bias
\begin{equation} \label{sudden} 
  p_1(\alpha) = \left\{ 
  \begin{array}{ll} 
     1 -p_{max} & \mbox{~for~} \alpha \leq \alpha_o. \\
     p_{max}>1/2 & \mbox{~for~} \alpha > \alpha_o.
     \end{array} \right. 
 \end{equation} 
We consider $p_{max}=0.75$ as an example, 
the corresponding results from the integration
of ODE and Monte Carlo simulations are shown
in Fig.\ \ref{suddenplot} for
training without weight decay (upper)
and for $\gamma=0.05$ (lower panel). 

We observe similar effects as for the slow,
linear time-dependence: The system reacts
rapidly with respect to the class-wise errors 
and the tracking error $\epsilon_{track}$ 
maintains a relatively low value. Also, the
reference error $\epsilon_{ref}$ displays 
robustness with respect to the sudden change
of $p_1$.  Weight decay, 
as can be seen in the lower panel of 
Fig.~\ref{suddenplot} reduces the over-all
sensitivity to the bias and its change: 
Class-wise errors are more balanced and the
weighted $\epsilon_{track}$ slightly increases
compared to the setting with $\gamma=0$.

\subsubsection{Periodic time dependence} 
\begin{figure}[t!]
\includegraphics[width=0.48\textwidth]{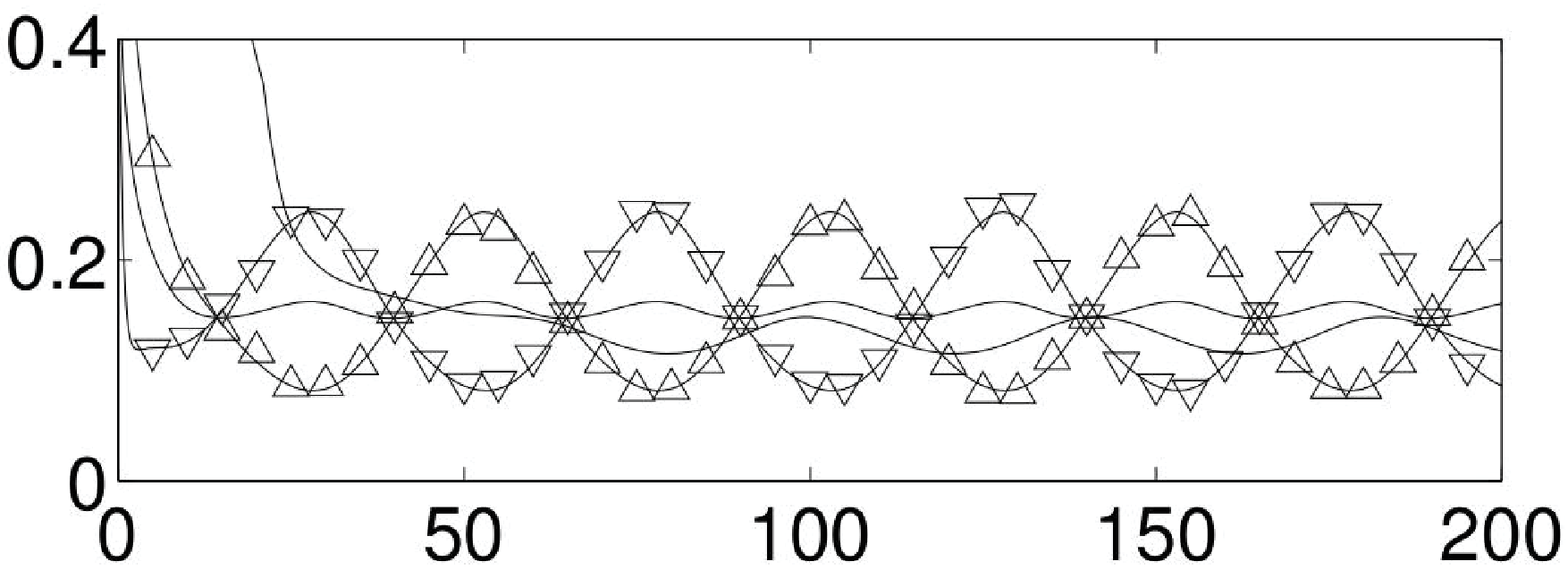}
\put(-90,0){\Large $\alpha$}
\put(-240,60){\Large  $\epsilon$}
\ \\ 
\includegraphics[width=0.48\textwidth]{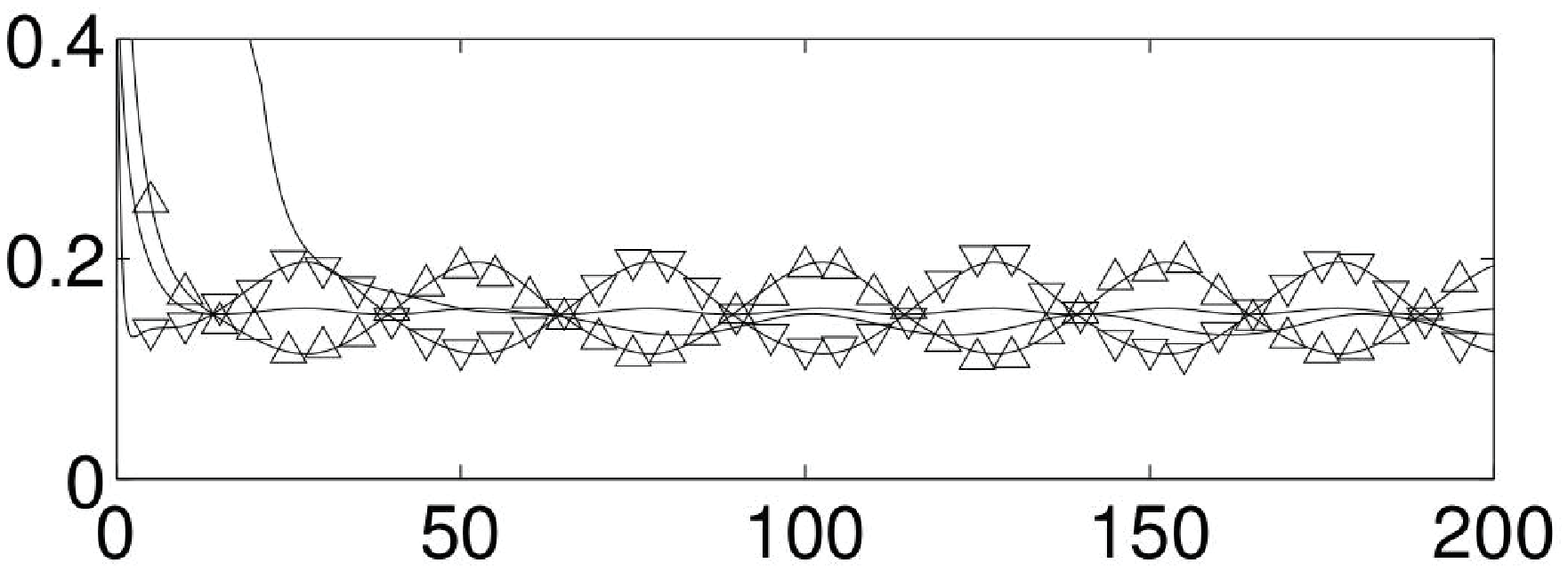}
\put(-90,0){\Large $\alpha$}
\put(-240,60){\Large  $\epsilon$}
\caption{ \label{periodicplot} 
LVQ1 in the presence
of oscillating class 
weights according
to Eq.\ (\ref{oscillating}) with parameters
$T=50$  and $p_{max}=0.8$, 
without weight decay $\gamma=0$ (upper graph)
and for $\gamma=0.05$ (lower). 
For clarity, Monte Carlo results are only shown
for the class-conditional errors $\epsilon^1$ 
(downward) and $\epsilon^2$ (upward triangles). 
All other details are given in  Fig.\  \ref{linplot}.} 
\end{figure}

As a third scenario we consider oscillatory
modulations of the class weights during training:  
\begin{equation} \label{oscillating}  
  p_1(\alpha) = 	 1/2 +\left (p_{max}-1/2\right) \, \cos\left( 2\pi \, 
  {\alpha}\big/{T} \right)
 \end{equation} 
 with periodicity $T$ on $\alpha$-scale and
 maximum amplitude $p_{max} <1$.
 Example results are shown in Fig.\ \ref{periodicplot} for $T=50$ and 
 $p_{max}=0.8$. Monte Carlo results for $N=100$
 are only displayed for the class-wise errors, for
 the sake of clarity. They 
 show excellent agreement with the 
 numerical integration of the ODE for training
 without weight decay (upper panel) and for 
 $\gamma=0.05$ (lower panel).
 These results confirm our findings for
 slow and sudden changes of the prior weights: 
 Weight decay limits the flexibility  of the LVQ
 system to react to the presence of a bias and its
 time-dependence. 
 
\subsubsection{Discussion: LVQ under virtual drift} 

Our results for the different realizations of time-dependent
class weights show that Learning Vector quantization can cope with this
form of drift to a certain effect. 
By design, standard incremental updates like the classical LVQ1
allow the prototypes to adjust to the changing statistics of the data.
This has been shown in \cite{entropy} for the actual drift of the cluster
centers in the model density.  Here we show that LVQ1 can also cope with
the virtual drift processes. 

In analogy to our findings in \cite{entropy}, one might have expected improved performance
when introducing weight decay as a mechanism of \textit{forgetting.}
As we demonstrate, however, weight decay does not have a very
strong effect on the the system's reaction to changing prior class weights. 
Essentially, weight decay limits the prototype norms and hinders
shifts of the decision boundary by prototype displacement. 
 The overall influence of class bias and its 
 time-dependence is reduced in the presence of weight decay. 
 Weight decay restricts the norm of the 
 prototypes, i.e. the possible offset of the decision boundary from the origin. 
 As a consequence, the tracking error slightly increases for $\gamma>0$, in general.  
 On the contrary, the error $\epsilon_{ref}$ with respect to the reference density 
 decreases compared to the training without weight decay. 

 A clear beneficial effect of \textit{forgetting} previous information
 in favor of the most recent examples cannot be 
 confirmed. The reaction of the learning system to sudden (B) or
 oscillatory changes of the priors (C) remains also unchanged when 
 introducing weight decay.


\subsection{Results: SCM regression under real drift}
\begin{figure*}[t]
\subfloat[]{\label{fig:erf_eg_curves_drift} \centering \includegraphics{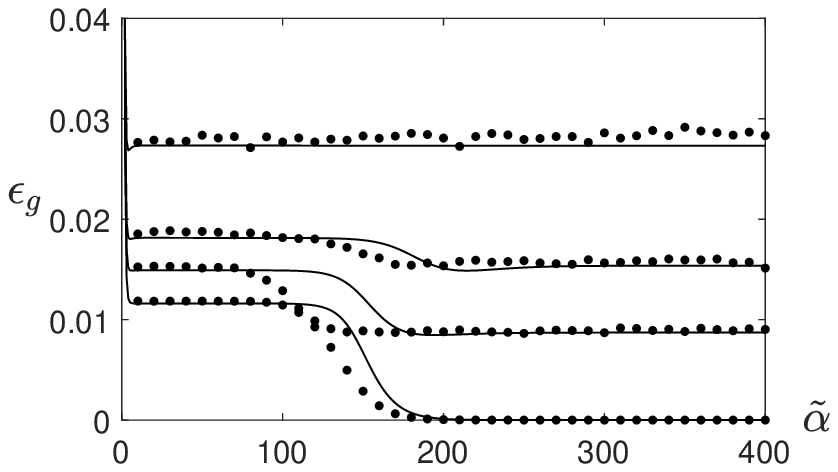}}
\subfloat[]{\label{fig:relu_eg_curves_drift} \centering \includegraphics{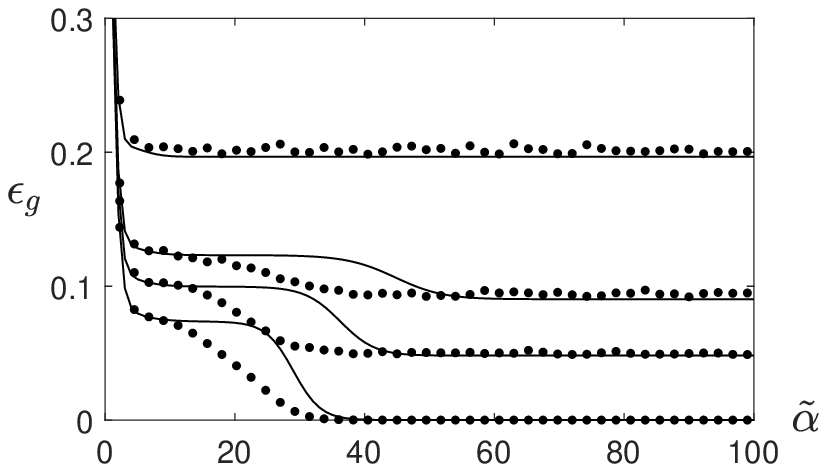}}
	\caption{The learning performance under concept drift in terms of generalization error as a function of the learning time $\widetilde{\alpha}$. Dots correspond to 10 runs of Monte Carlo simulations with $N=500$, $\eta=0.05$ with initials conditions as in Eq.~(\ref{eq:SCMInitOrderParams}). Solid lines show ODE integrations. (a): Erf SCM. From bottom to top, the curves correspond to the levels of target drift $\tilde{\delta}=\{0,0.01,0.02,0.05\}$. (b): ReLU SCM. From bottom to top, the levels of target drift are: $\tilde{\delta}=\{0,0.05,0.1,0.3\}$.}\label{fig:eg_curves_drift}
\end{figure*}

Here we present the results concerning the SCM student teacher scenario with $K=M=2$ 
under real concept drift, i.e.\ random displacements of the teacher
vectors as introduced in Sec.\ \ref{realdrift}. 
{Unlike LVQ for classification,
gradient descent based training of a regression system
is expected to be much more sensitive
to the choice of the learning rate. Here, we restricted the discussion to the well-defined limit of 
small learning rates, $\eta\to 0$ and $\alpha\to \infty$
with $\tilde{\alpha} = \eta\alpha = {\cal O}(1),$ see the 
discussion before Eq. (\ref{alphatilde}). In the 
corresponding Monte Carlo simulations, cf. Figs.\ \ref{fig:erf_eg_curves_drift}
and 
\ref{fig:relu_eg_curves_drift}, 
we employed a small learning rate $\eta=0.05$ which 
yielded very good agreement.}

Already in the absence of concept drift, the  model
displays non-trivial effects as shown in, for instance,
\cite{transient,saadsolla1,saadsolla2}.
Perhaps the most thoroughly studied phenomenon in the SCM training process 
is the existence of quasi-stationary plateaus in the evolution of the order
parameters and the generalization error. 
In the most clear-cut cases, they correspond to approximately symmetric 
configurations of the student network with respect to the teacher network, 
i.e. $R_{im} \approx R$ for all $i,m.$ In such a state, all
student units have acquired the same, limited knowledge of the 
target rule. Hence, the generalization error in the plateau is sub-optimal. In terms of 
Eqs. (\ref{odegeneric}),   
plateaus correspond to weakly repulsive fixed points of the ODE system. One can show
in case of orthonormal 
teacher units and for small learning rates that a  symmetric 
fixed point with $R_{im}=R$ and the associated plateau state
always exists, see e.g.\ \cite{saadsolla2}.  
In order to achieve a further decrease of the generalization error, the symmetry of the
student with respect to the teacher units has to be broken by \emph{specialization}: Each 
student weight vector $\vec{w}_{1,2}$ has to represent a specific teacher 
unit and $R_{i1} \neq R_{i2}$ is required for successful learning.

Our recent comparison of Erf-SCM and ReLU-SCM  
revealed interesting differences even in absence of concept drift
\cite{ReLUesann}. For instance, 
in the Erf-SCM,  student vectors are nearly identical
in the symmetric plateau with $Q_{ik} \approx 
Q$ for all $i,k \in \{1,2\}.$ On the contrary, 
in ReLU systems the
student weights are 
not aligned in the quasi-stationary state: 
$Q_{ii}=Q$ and $Q_{12}<Q$ \cite{ReLUesann}. 

\subsubsection{ODE and Monte Carlo simulations}  \label{ODEMC}

Here, we investigate and compare the learning dynamics of 
networks with Erf- and ReLU-activation under concept drift and 
in the presence of weight decay. To this 
end we study the models by numerical integration of the corresponding ODE and, 
in addition, by Monte Carlo simulations. 

We study  training processes in absence of prior knowledge in the student.
In the following we consider exemplary initial conditions with 
\begin{equation}\label{eq:SCMInitOrderParams}
R_{im}(0)=0, \> Q_{11}(0)=Q_{22}(0)=0.5, \> Q_{12}(0)=0.49\, 
\end{equation}
which correspond to almost identical $\mathbf{w}_1(0)$ and $\mathbf{w}_2(0),$ which
are both orthogonal to the teacher vectors. Note that the initial norm of the student 
vectors and their mutual \textit{overlap} $Q_{12}(0)$ can
be set arbitrarily in practice. 

For the networks with two hidden units
we define the quantity $S_i(\alpha)=|R_{i1}(\alpha) - 
R_{i2}(\alpha)|$ as the specialization of student units $i=1,2$. In the plateau 
state, $S_i(\alpha) \approx 0$ for an extended amount of training time, while
an increasing value of 
$S_i(\alpha)$ indicates the specialization of the unit.
In practice, one expects that initially $R_{im}(0) \approx 0$ for all $i,m$ if no prior 
information is available about the target rule. Hence, the student specialization $S_i(0) = 
|R_{i1}(0) - R_{i2}(0)|$ is also small, initially. 

The unspecialized plateau can dominate the learning process and, consequently, its length
is a quantity of significant interest. 
Quite generally, it is governed by the  repulsive properties of the relevant fixed 
point of the ODE system and depends
logarithmically on the the magnitude of the initial 
specialization $S_i(0)$, see \cite{transient} for a detailed discussion. 
In simulations for large $N$, a random initialization of student vectors 
would result in overlaps $R_{im}(0)={\cal O}(1/\sqrt{N})$ with the teacher vectors
which also implies that $S_i(0)={\cal O}(1/\sqrt{N}).$
The accurate extrapolation of simulation results for $N\to\infty$ is 
complicated by this interplay of finite size effects and initial specialization 
which governs the escape from the plateau states \cite{transient}. Due to fluctuations
in a finite system, plateaus are typically left earlier than predicted by the theoretical
prediction for $N\to\infty$. 
Here we focus on the performance achieved in the plateau states and
resort to a simpler strategy: 
The values of the order parameters observed at $\widetilde{\alpha}=0.05$ in the
Monte Carlo simulation are used as initial 
values for the numerical integration of the ODE.
This does not necessarily warrant a one-to-one
correspondence of the precise shape and length of the 
plateau states. However, the comparison shows excellent qualitative agreement 
and allows for the quantitative comparison of the performance in the quasistationary
and states. 

We have studied the Erf-SCM and the ReLU-SCM under concept drift, 
Eq.~(\ref{Bdrift}), and weight decay, Eq.~(\ref{withdecay}), in 
the limit of small learning rates $\eta \to 0$. We  resorted to this simplifying limit 
as the term $G_{ik}^{(2)}$ in Eq.\ (\ref{FGSCM}) could not be obtained 
analytically for the ReLU-SCM. However, non-trivial results can be achieved
in terms of the rescaled training time $\tilde{\alpha}$ 
in the limit (\ref{alphatilde}). 
Hence we integrate the ODE 
provided in Eq.~(\ref{odesmallrate}),  combined with
the drift and weight decay terms from Eqs.~(\ref{odedrift},\ref{odedecay}) that also have to be scaled with $\eta$ in this case: $\tilde{\delta} = \eta \delta$, $\tilde{\gamma} = \eta \gamma$.
In addition to 
the numerical integration  we have performed and
averaged over 10 independent runs of Monte Carlo simulations with 
system size $N=500$ and small but finite learning rate $\eta=0.05$.

\subsubsection{Learning curves under concept drift} 

Fig.~\ref{fig:eg_curves_drift} shows the learning curves $\epsilon_g (\tilde{\alpha})$ 
as results of
the averaged Monte Carlo simulations and the ODE integration for different 
strengths $\tilde{\delta}$ of concept drift with no weight decay
($\tilde{\gamma}=0$). The left and right panel corresponds to Erf- and ReLU-SCM, respectively. 

Apart from deviations in terms of the plateau lengths, 
simulations and  the numerical integration of the ODE show very
good agreement. In particular, the generalization error in the 
plateau and  final states nearly coincides. As outlined in Sec.\ \ref{ODEMC},
the actual length of plateaus in simulations depends on subtle
details \cite{transient} which were not addressed here. 

Note also that a direct, quantitative comparison of Erf- and ReLU-SCM in terms of
training times $\tilde{\alpha}$ is not meaningful. For instance, it seems 
tempting to conclude that the ReLU-SCM exhibit shorter plateau states for
the same network size and training conditions. However, one has 
to take into account that the activation
functions influence the complexity of the 
input output relation of the network in a non-trivial way.  

From the behavior of the learning curves  for increasing  strengths 
$\tilde{\delta}$, several 
impeding effects of the drift can be identified: The generalization error in the 
unspecialized plateau and in the final state for large $\tilde{\alpha}$
increase with $\tilde{\delta}$. 
At the same time, the plateau lengths increase. These effects are observed
for both types of activation function. 
More specifically, the behavior for small $\tilde{\delta}$ is close to the stationary setting
with $\tilde{\delta}=0$: A rapid initial decrease of the generalization error is followed by the 
quasi-stationary plateau state that persists for a relatively long training time.
Eventually, the system escapes from the plateau and improved generalization performance 
becomes possible. Despite the matching complexity of student and teacher,  
perfect generalization cannot be achieved in the presence of on-going concept drift. 

We note that the stronger the drift, the smaller is the difference between
the performance in the plateau and the final state. 
For very large values of $\tilde{\delta}$ both versions of the SCM cannot escape the 
plateau state anymore as it corresponds to a stable fixed point
of the ODE. 

In the following we discuss
for both activation functions the effect of concept drift on the
plateau- and final generalization error in greater detail. The influence of 
weight decay on the dynamics is also presented. 

\begin{figure*}[t]
\subfloat[]{\label{fig:erf_egfp_drift}
            \includegraphics[width=0.47\textwidth]{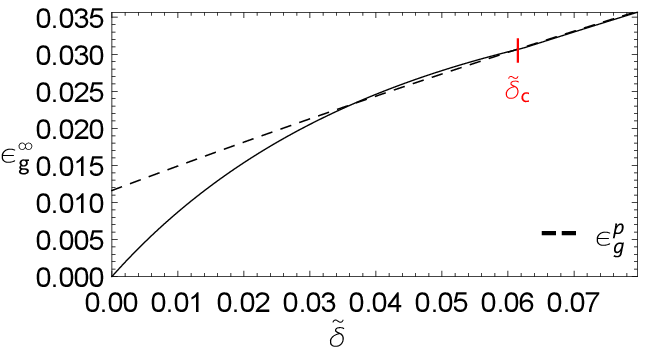}} \quad
\subfloat[]{\label{fig:erf_egfp_wd}
            \includegraphics[width=0.47\textwidth]{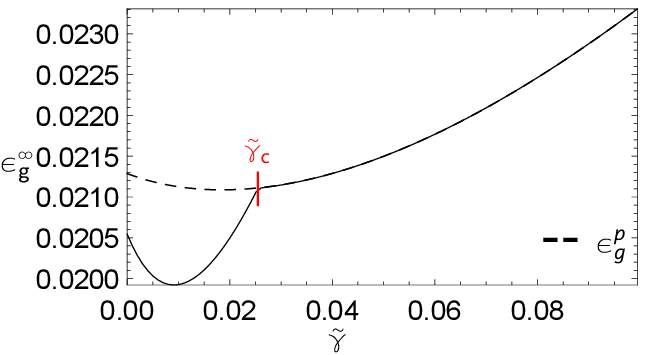}}\\[-0.1ex]
\subfloat[]{\label{fig:erf_pl_drift}
            \includegraphics[width=0.47\textwidth]{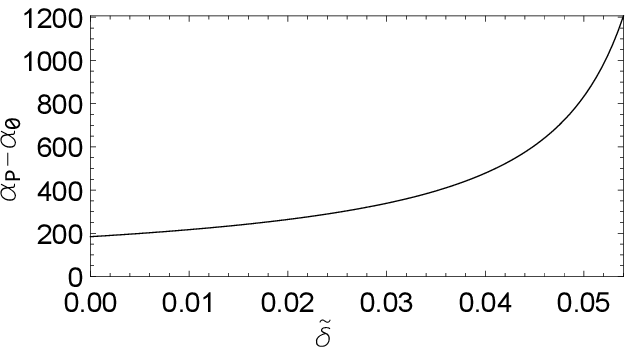}} \qquad
\subfloat[]{\label{fig:erf_pl_wd}
            \includegraphics[width=0.47\textwidth]{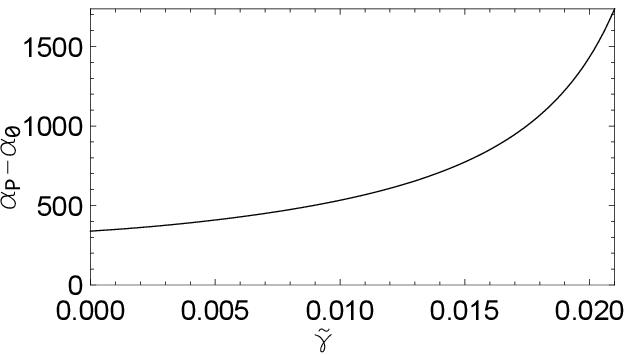}}
	\caption{Erf-SCM: Generalization error under concept drift in unspecialized plateau states (dashed lines) and final states (solid) of the learning process.   \ref{fig:erf_egfp_drift}: Plateau- and final generalization error for an increasing 
	strength $\tilde{\delta}$ of the target drift. Here, weight decay is not applied: $\tilde{\gamma}=0$.
    For $\tilde{\delta}>\tilde{\delta}_c$ as marked by the vertical line, the curves merge. 
	\ref{fig:erf_egfp_wd}: The plateau- and final generalization error as 
	a function of the weight decay parameter $\tilde{\gamma}$
	for a fixed level of real target drift, here: $\tilde{\delta}=0.03$. The curves merge for $\tilde{\gamma}>\tilde{\gamma}_c$, as marked by the vertical line.
	The lower panels show the observed plateau lengths as a function of $\tilde{\delta}$
	for $\tilde{\gamma}=0$ (5c) and as a function of $\tilde{\gamma}$ for fixed $\tilde{\delta}=0.03$
	(5d), respectively. 
	}\label{fig:erf_drift_results}
\end{figure*}

\paragraph{Erf-SCM under drift and weight decay} \ \\ 
\noindent
Fig.~\ref{fig:erf_egfp_drift} displays the effect of the drift strength $\tilde{\delta}$  
on the generalization error in the unspecialized plateau state and in the final 
state for $\tilde{\alpha}\to\infty$,
i.e. $\epsilon_g^p(\tilde{\delta})$ and $\epsilon_g^\infty(\tilde{\delta}),$ respectively. As mentioned above, weak drifts 
still allow for student specialization with improved performance in the final state for
large $\tilde{\alpha}$. However, increasing the drift strength results
in a decrease of the difference 
\(|\epsilon_g^\infty(\tilde{\delta}) - \epsilon_g^p(\tilde{\delta})|.\)
We have marked the value of $\tilde{\delta}$, above which the plateau becomes the stable 
final state for $\tilde{\alpha}\to\infty$
in the figure and 
refer to it as $\tilde{\delta}_c$.

Interestingly, in a small range of the drift parameter, $0.036 < \tilde{\delta} < 0.061$,  
the final performance is actually worse than in the plateau with $\epsilon_g^\infty(\tilde{\delta}) > 
\epsilon_g^p(\tilde{\delta})$. Since $\epsilon_g$ depends explicitly also  
on the $Q_{ik}$, it is possible for an unspecialized state with $R_{im}=R$ to
generalize better than a slightly specialized configuration with unfavorable values
of the student norms and mutual overlaps. 

Fig.~\ref{fig:erf_pl_drift} shows the effect of the drift on the plateau length. The start and end of the plateau are defined as
\begin{eqnarray}
    \tilde{\alpha}_0 &=& \min \{ \tilde{\alpha} \, | \, \epsilon_g^p - 10^{-4} < \epsilon_g(\tilde{\alpha}) < \epsilon_g^p + 10^{-4} \} \nonumber \ \\
    \tilde{\alpha}_P &=& \min \{ \tilde{\alpha} \, | \, S_i(\tilde{\alpha}) \geq 0.2 \, S_i(\tilde{\alpha} \to \infty) \} \, .
\end{eqnarray}
Here, $S_i(\tilde{\alpha}\to\infty)$ represents the  final specialization that is achieved 
by the system for large training times. $(\alpha_P - \alpha_0)$ is used as a meaure of the plateau length.

In the weak drift regime, the plateau length increases slowly with $\tilde{\delta}$
as shown in panel (c) for $\tilde{\gamma}=0$. It eventually 
diverges as $\tilde{\delta}$ approaches $\tilde{\delta}_c$  from 
Fig.~\ref{fig:erf_egfp_drift}.

The dependence of $\epsilon_g^p$ and $\epsilon_g^\infty$ on the weight decay parameter $\tilde{\gamma}$ is shown in
Fig.~\ref{fig:erf_egfp_wd}. We observe improved performance for a small amount weight 
decay compared to absence of weight decay ($\tilde{\gamma} = 0$). However, the system is quite
sensitive to the actual setting of $\tilde{\gamma}$: Values slightly larger than the optimum quickly deteriorate the ability for improvement from the plateau generalization error. The value of $\tilde{\gamma}$, above which the plateau- and final generalization error coincide has been marked in the figure and we refer to it as $\tilde{\gamma}_c$.

Fig.~\ref{fig:erf_pl_wd} shows the effect of the weight decay on the plateau length in the 
same setting as in Fig.~\ref{fig:erf_egfp_wd}. Introducing a weight decay always extends the plateau length. For small $\tilde{\gamma}$ the plateau length grows
slowly and diverges as $\tilde{\gamma}$ approaches $\tilde{\gamma}_c$ from Fig.~\ref{fig:erf_egfp_wd}.

\begin{figure*}[t]
\subfloat[]{\label{fig:relu_egfp_drift}
            \includegraphics[width=0.48\textwidth]{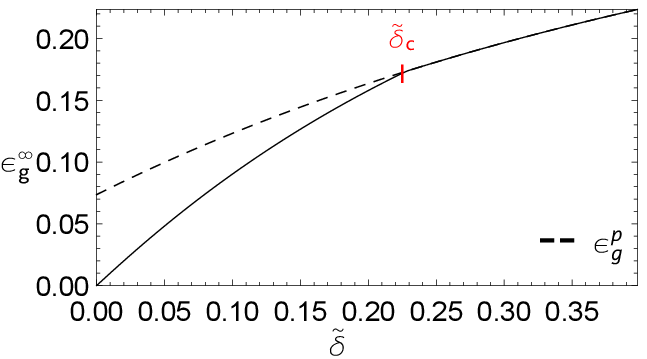}} \quad
\subfloat[]{\label{fig:relu_egfp_wd}
            \includegraphics[width=0.48\textwidth]{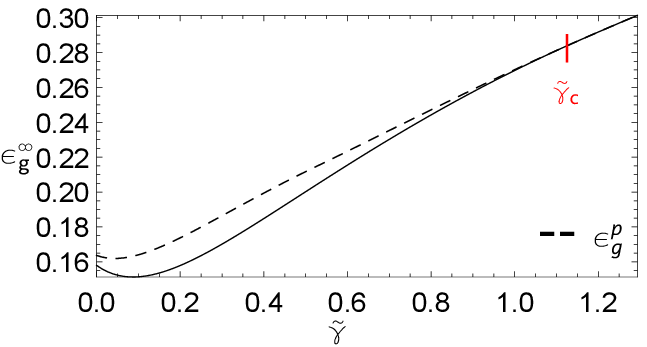}}\\[-2ex]
            
\subfloat[]{\label{fig:relu_pl_drift}
            \includegraphics[width=0.48\textwidth]{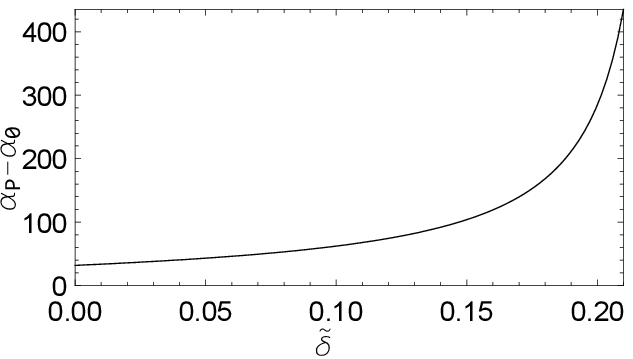}} \quad
\subfloat[]{\label{fig:relu_pl_wd}
            \includegraphics[width=0.48\textwidth]{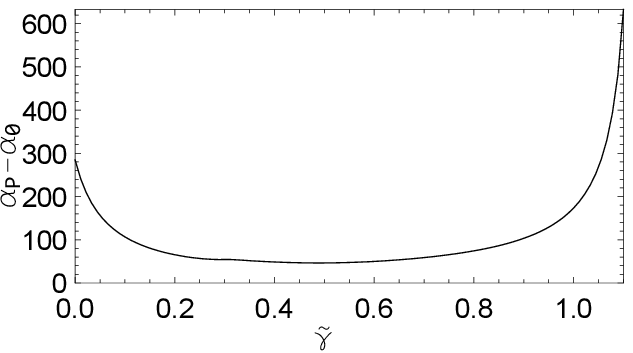}}
	\caption{ReLU-SCM: Generalization error under concept drift in unspecialized plateau states (dashed lines) and final states (solid), as a function of the drift strength (6a) and weight decay (6b). In (6b), $\tilde{\delta}=0.2$. The drift strength $\tilde{\delta}_c$ above which the curves merge is marked in (6a) and similar for weight decay $\tilde{\gamma}_c$ in (6b). The lower panels show the observed plateau lengths as a function of $\tilde{\delta}$
	for $\tilde{\gamma}=0$ (6c) and as a function of $\tilde{\gamma}$ for fixed $\tilde{\delta}=0.2$
	(6d), respectively.}\label{fig:relu_drift_results}
\end{figure*}


\paragraph{ReLU-SCM under drift and weight decay} \ \\
\noindent
The effect of the strength of the drift on the generalization error in the unspecialized plateau state and in the final state is 
displayed in Fig.~\ref{fig:relu_egfp_drift}. The picture is similar to the Erf-SCM: an increase in the drift strength causes an increase in the plateau- and final generalization error. We have marked in the figure the drift strength $\tilde{\delta}_c $ at which there is no further change in performance from the plateau. In contrast to the Erf-SCM,
there is no range of $\tilde{\gamma}$ for which the ReLU-SCM generalization error increases
after leaving the plateau.

Fig.~\ref{fig:relu_pl_drift} shows the effect of the strength of the drift on the plateau length. Here too, a similar dependence is observed as for the Erf-SCM: For the range of weaker drifts, the plateau length grows slowly and diverges for strong drifts up to the drift strength $\tilde{\delta}_c$ from Fig.~\ref{fig:relu_egfp_drift}.

Fig.~\ref{fig:relu_egfp_wd} shows the effect of the amount of weight decay on the plateau- and final generalization error in a concept drift situation. A small amount of weight decay can improve the generalization error compared to no weight decay ($\tilde{\gamma}=0$). The effect weight decay has on the ReLU-SCM shows a much greater robustness compared to the Erf-SCM in terms of the ability to improve from the plateau value: For high amounts of weight decay, an escape from the plateau to better performance can still be observed. The value $\tilde{\gamma}_c$, above which the plateau- and final generalization error coincide has been marked in the figure.

Fig.~\ref{fig:relu_pl_wd} shows the effect of the amount of weight decay on the plateau length in the same concept drift situation as in Fig.~\ref{fig:relu_egfp_wd}. It shows that the plateau is shortened significantly in the smaller range of weight decay, the same range that also improves the final generalization error as observed in Fig.~\ref{fig:relu_egfp_wd}. The plateau length increases again for very high levels of weight decay and diverges as $\tilde{\gamma}$ approaches the $\tilde{\gamma}_c$ from Fig.~\ref{fig:relu_egfp_wd}. 

\subsection{Discussion: SCM regression under real drift}
As was already discussed, the symmetric plateau corresponds to states where the student units have all learned the same limited and general knowledge about the teacher units, i.e. $R_{ij} \approx R$ and therefore the specialization of each student unit $i$ is small: $S_i(\tilde{\alpha}) = |R_{i1}(\tilde{\alpha}) - R_{i2}(\tilde{\alpha})| \approx 0$. Eventually, the symmetry is broken by the start of specialization, when $S_i(\tilde{\alpha})$ increases for each student unit $i$. For stationary learnable situations with $K=M$, throughout learning the students units will acquire a full overlap to the teacher units: $S_i = 1$ for all student units $i$. In this configuration, the target rule has been fullly learned and therefore the generalization error is zero. 
In our modelled concept drift, the teacher vectors are changing continuously. This reduces the overlaps the student units can achieve with the teacher units, which increases the generalization error in the plateau state and the final state.

Identifying the specific teacher vectors is more difficult than learning the general structure of 
the teacher: Hence, increasing the drift causes the final generalization error to deteriorate faster 
than the plateau generalization error. For very strong target drift, the teacher vectors are 
changing too fast for specialization to be possible. We have identified the strength of the drift 
above which any kind of specialization is impossible for both SCM by studying the properties of the 
fixed point in the ODE. In stationary situations, one eigenvalue of the linearized dynamics near the 
fixed point is positive and causes the repulsion away from the fixed point to specialization. We 
refer to this positive eigenvalue as $\lambda_s$. 
The eigenvalue decreases linearly with the drift strength:
For small $\tilde{\delta}$, $\lambda_s$ is still positive 
and therefore an escape from the plateau is observed. However, $\lambda_s$
is negative for $\tilde{\delta} > \tilde{\delta}_c$, the symmetric fixed 
point is stable and specialization becomes impossible. For the 
Erf-SCM, $\tilde{\delta}_c \approx 0.0615$ and for the ReLU-SCM $\tilde{\delta}_c \approx 0.225$.
The weaker repulsion of the fixed point for stronger drift causes the plateau length to grow for $\tilde{\delta} \to \tilde{\delta}_c$. In practice, this implies that higher training effort is necessary the stronger the concept drift is.

In the $\tilde{\alpha}\to\infty$ final state, the student tracks the drifting target rule. For 
$\tilde{\delta} \ll \tilde{\delta}_c$, the student can achieve highly specialized states while tracking the teacher. The closer the drift strength is to $\tilde{\delta}_c$, the weaker is the specialization that can be achieved by the student while following the rapidly moving teacher vectors. For $\tilde{\delta} > \tilde{\delta}_c$, 
the unspecialized student can only track the rule in terms of a simple approximation.

In the results of the Erf-SCM, a range of drift strength $0.036 < \tilde{\delta} < \tilde{\delta}_c$ was observed for which the final generalization error in the tracking state is worse than the plateau generalization error. Upon further inspection, this is due to the large values of $Q_{11}$ and $Q_{22}$ of the student vectors in the specialized regime. Hence, the effect can be prevented by 
introducing an appropriate weight decay.

\subsubsection{Erf SCM vs. ReLU SCM: Weight decay in concept drift situations}
Our results show that weight decay can improve the final generalization error in the specialized tracking state for both SCM. The suppression of the contributions of older and thus less representative data shows benefits in both systems.

However, from the result in Fig.~\ref{fig:erf_egfp_wd}, we find that it is particularly important to 
tune the weight decay parameter for the Erf-SCM, since the specialization ability deteriorates 
quickly for values slightly off the optimum, as shown in the figure by the rapid increase in 
$\epsilon_g^\infty$. This reflects a steep decrease of the largest
eigenvalue $\lambda_s$ in the ODE for 
the Erf-SCM with $\tilde{\gamma}$, which also causes the increase of the plateau length as
observed in Fig.~\ref{fig:erf_pl_wd}. Already from $\tilde{\gamma}_c \approx 0.0255$, the eigenvalue $\lambda_s$ becomes negative, and 
therefore the fixed point becomes an attractor.

We found a very different effect of weight decay on the performance of the ReLU-SCM. Not only is it able to improve the final generalization error in the tracking state as shown in Fig.~\ref{fig:relu_egfp_wd}, but it also significantly reduces the plateau length in the lower range of weight decay. This reflects the increase of $\lambda_s$ with the weight decay parameter in the fixed point of the ODE, which increases the repulsion from the unspecialized fixed point. 
Clearly, suppressing the contribution of older data is beneficial for the specialization ability of the ReLU-SCM. For larger values of $\tilde{\gamma},$ the plateau length increases, reflecting a decrease of $\lambda_s$. However, specialization remains possible up to a rather high value of weight decay $\tilde{\gamma}_c \approx 1.125$. The greater robustness to weight decay with respect to specialization as shown in Fig.~\ref{fig:relu_egfp_wd} is likely related to our previous findings in \cite{ReLUesann}, which show that the ReLU student-teacher setup needs fewer examples to reach specialization. We hypothesize that the simple linear nature of the function makes it easier for the student to learn features of the target rule. Hence a relatively small window of recent examples can already facilitate a degree of specialization.

\section{Summary and Outlook} 
\vspace{-2mm}
We have presented
a  mathematical framework
in which to study the influence of concept drift
systematically in model scenarios. We exemplified the
use of the versatile approach in terms of models for the
training of 
prototype-based classifiers (LVQ) and shallow neural networks for
regression, respectively. 

\paragraph{LVQ for classification under drift and weight decay}  \ \\
\noindent 
In all specific drift scenarios considered here,
we observe that
simple LVQ training can track the time-varying class
bias to a non-trivial extent:
In the
interpretation of the results in terms
of real drift, the class-conditional performance
and the 
tracking error $\epsilon_{track}(\alpha)$ clearly 
reflect the time-dependence of the prior weights.
In general, the reference error 
$\epsilon_{ref}(\alpha)$ 
with respect to  class-balanced test data,  
displays only little
deterioration due to the drift in the
training data.
The main effect of introducing weight decay 
is  a reduced overall sensitivity to bias in the
training data: Figs.\ 1-3 display
a decreased difference between the class-wise
errors $\epsilon^{1,2}$ for 
$\gamma>0$. 
Na\"ively, one might have expected
an improved tracking of the drift due to the
imposed \textit{forgetting}, resulting in, for instance, 
a more rapid reaction to the 
sudden change of bias in Eq.\  (\ref{sudden}).  
However, such
an improvement cannot be confirmed. 
This finding is in contrast to a recent study
\cite{entropy}, in which we observe    
increased performance by weight
decay for a particular real drift, i.e.\ 
the randomized displacement of
cluster centers. 

The precise
influence of weight decay clearly 
depends on the
geometry and relative position of the clusters.
Its dominant effect, however, is the 
regularization
of the LVQ system by 
reducing the norms 
of the prototype vectors.
Consequently, the NPC classifier
is less flexible to reflect 
class bias which would require
significant offset
of the prototypes and decision boundary 
from the origin. This mildens the influence
of the bias and its time-dependence and it
results in a more robust behavior of the
employed error measures.

\paragraph{SCM for regression under drift and weight decay} \ \\
\noindent
On-line gradient descent learning in the SCM has proven able to cope with drifting concepts
in regression: 
For weak drifts, the SCM still achieves significant specialization
with respect to the drifting teacher vectors, although the required learning time 
increases with the strength of the drift.
In practice, this results in higher training effort to reach beneficial states in the 
network. The drift constantly reduces the overlaps with the teacher vectors which 
deteriorates the performance. After reaching a specialized state, the network efficiently
tracks the drifting target. However, in the presence of very strong drift,
both versions of the SCM (with Erf- and ReLU-activation) 
lose their ability to specialize and as a consequence 
their generalization behavior remains poor.

We have shown that weight decay can improve the performance in the plateau and in the  
final tracking state. For the Erf-SCM, we found that there is a small range in which weight 
decay yields favorable performance while the network quickly loses the specialization 
ability for values outside this range. Therefore, in practice a careful tuning of the weight 
decay parameter would be required. The ReLU network showed greater robustness to the 
magnitude of the weight decay parameter and displayed a stronger tendency to specialize. 
Weight decay also reduced the plateau length significantly in the training of ReLU 
SCM. Hence, weight decay could speed up the training of ReLU networks in practical concept 
drift situations, achieving favorable weight configurations more efficiently. 
Also, the network  performs well with a larger range of the weight decay parameter and does not 
require the careful tuning necessary for the Erf-SCM.


\paragraph{Outlook}  \ \\
\noindent
{The presented modelling framework offers the possibility
to extend the scope of our studies in several relevant directions. 
For instance, 
the formalism facilitates the consideration of more complex model scenarios.
Greater values of $K$ and $M$ should be studied in, both, classification and regression.
While we expect key results to carry over from $K=M=2$, 
the greater complexity of the systems should result in 
richer dynamical behavior in detail. 
We will study if and how a mismatched number of prototypes further impedes
the ability of LVQ systems to react appropriately to the presence of concept drift. 
The training of an SCM with $K\neq M$ should be of considerable interest and 
will also be addressed in forthcoming studies. 
One might speculate that concept drift could enhance 
overfitting effects in over-sophisticated SCM with $K>M$ hidden units.  Ultimately, the characteristic robustness of the ReLU 
activation function to weight decay that was found should be studied in practical 
situations. Qualitative 
results are likely to carry over to similarly shaped activation functions, which will be
verified in future work.

In a sense, the considered sigmoidal and ReLU activation functions are prototypical
representatives of the most popular choices in machine learning practice. 
The extension to various 
modifications or significantly different transfer functions \cite{timetoswish,deep}
should provide additional valuable insights of practical relevance. 
Exact solutions to the averages that are necessary for the formulation of the learning 
dynamics in the thermodynamic limit may not be available for all activation functions. 
In such cases we can resort to approximations schemes and simulations.

The consideration of more complex input densities will also shed 
light on the practical
relevance of our theoretical investigations. Recent work \cite{Goldt,capturing} 
shows that the 
statistical physics based investigation of machine learning processes can take into
account realistic input densities, bridging the gap between the theoretical models
and practical applications. } 

Our modeling framework can also be applied in the analysis of other types 
of drift or combinations thereof. 
Several virtual processes could readily be implemented in the model
of LVQ training: time-dependent characteristics of the input density could
include the variances of the clusters or their relative position in feature space. 
A number of extensions is also possible in the regression model. For instance,
teacher networks with time-dependent complexity
could be studied by varying the mutual teacher overlaps $\vec{B}_{m}\cdot\vec{B}_n$ 
in the course of training. 

Alternative mechanisms of \textit{forgetting} beyond weight decay should
be considered, which do not limit the flexibility
of the trained systems as drastically.
As one example strategy  we intend to investigate the accumulation of additive 
noise in the training processes.  We will also explore the parameter space of
the model density in greater depth and 
study the influence of the learning rate systematically.

One of the major challenges in the field 
is the reliable detection of concept drift in a stream of data. 
Learning systems should be able to discriminate 
drift from static noise in the data and infer  
also the type of drift, e.g. virtual vs. real.   
Moreover, the strength of the drift has to be estimated
reliably in order to adjust the
training prescription accordingly.  It could be highly
beneficial to extend our framework towards efficient drift detection and estimation 
procedures.

\begin{acknowledgements}
{M.B. and M.S. acknowledge support through the \textit{Northern Netherlands Region of Smart Factories (RoSF)} consortium, lead by the 
\textit{Noordelijke Ontwikkelings en Investerings Maatschappij (NOM)}, 
The Netherlands, see {\tt{http://www.rosf.nl}}.
B.H. gratefully acknowledges funding by \textit{Bundesministerium f{\"ur} Bildung und Forschung} (BMBF) under grant number 01IS18041A.
}
\end{acknowledgements}

\section*{Conflict of interest} The authors declare that they have no conflict of interest.


\bibliographystyle{spbasic}      

\bibliography{biblio}   


\end{document}